\pdfoutput=1

\documentclass[final]{clv2025}

\pageonefooter{Action editor: Makoto Miwa. Submission received: October 21, 2025; revised version received: February 20, 2026; accepted for publication: March 28, 2026.}

\usepackage{amsmath,amssymb}
\usepackage{graphicx}
\usepackage{booktabs}
\usepackage{url}

\usepackage{latexsym}

\usepackage[T1]{fontenc}

\usepackage[utf8]{inputenc}

\usepackage{microtype}

\usepackage{makecell}
\usepackage{enumitem}
\usepackage{multirow}
\usepackage{multicol}
\usepackage[most]{tcolorbox}
\usepackage{adjustbox}
\usepackage{subcaption}
\usepackage{soul}
\usepackage{float}
\usepackage[normalem]{ulem}
\useunder{\uline}{\ul}{}
\restylefloat{table}

\usepackage[dvipsnames]{xcolor}

\newcommand{\bestcomp}[1]{\underline{#1}}  
\newcommand{\bestsuff}[1]{\textbf{#1}}     

\newcommand{\subj}[1]{\textcolor{blue}{#1}}
\newcommand{\gold}[1]{\textcolor{orange}{#1}}
\newcommand{\noise}[1]{\textcolor{green}{#1}}

\usepackage{tikz}
\usetikzlibrary{shapes}
\usepackage{nicefrac}
\usepackage{amsfonts}

\newcommand{\rankk}{$Rank@k$}

\newcommand{\mrr}{$MRR$} 
\newcommand{\mrre}{MRR} 

\newcommand{\rankke}{Rank@k} 
\newcommand{\Drankke}{\Delta Rank@k}

\newcommand{\rankkeInst}{\rankke^{\mathrm{inst}}}
\newcommand{\rankkInst}{$\rankke^{\mathrm{inst}}$}

\newcommand{\DrankkeGroup}{\Drankke^{\mathrm{grp}}}
\newcommand{\DrankkeInst}{\Drankke^{\mathrm{inst}}}

\newcommand{\DrankkGroup}{$\Drankke^{\mathrm{grp}}$}
\newcommand{\DrankkInst}{$\Drankke^{\mathrm{inst}}$}

\usepackage{xspace}

\newcommand{\NMutInfAt}[1]{\ensuremath{\mathrm{NMutInf@}#1}\xspace}
\newcommand{\MDLBitsAt}[1]{\ensuremath{\mathrm{MDL\text{-}Bits@}#1}\xspace}

\newcommand{\NMutInfk}{\NMutInfAt{k}}
\newcommand{\MDLBitsk}{\MDLBitsAt{k}}

\newcommand{\MI}{MechLight} 
\newcommand{\FA}{FA} 
\newcommand{\IG}{IG} 
\newcommand{\Attn}{ATTN}

\usepackage{mdframed}
\usepackage{hyperref}
\hypersetup{hidelinks}

\AtBeginEnvironment{blockquote}{\small}
\definecolor{block-gray}{gray}{0.85}
\newtcolorbox{blockquote}{colback=block-gray,boxrule=0pt,boxsep=0pt,breakable}

\AtBeginEnvironment{definition}{\small}

\runningtitle{Evaluation Framework for Highlight Explanations of Context Utilisation in Language Models}
\runningauthor{Sun et al. }   

\title{Evaluation Framework for Highlight Explanations of Context Utilisation in Language Models}

\author{Jingyi Sun$^{1}$\thanks{Equal contribution.}\thanks{Corresponding author.} ,\;
        Pepa Atanasova$^{1}$\footnotemark[1] ,\;
        Sagnik Ray Choudhury$^{2}$ ,\;
        Sekh Mainul Islam$^{1}$,\;
        Isabelle Augenstein$^{1}$
        }

\affilblock{
  \affil{University of Copenhagen, Copenhagen, Denmark\\
  \texttt{\{jisu,pepa,seis,augenstein\}@di.ku.dk}}
  \affil{University of North Texas, Denton, Texas, USA\\
  \texttt{sagnik.raychoudhury@unt.edu}}
}

\begin{document}
\maketitle
\begin{abstract}

Context utilisation, the ability of Language Models (LMs) to incorporate relevant information from the provided context when generating responses, remains largely opaque to users, who cannot determine whether models draw from parametric memory or provided context, nor identify which specific context pieces inform the response. Highlight explanations (HEs) offer a natural solution as they can point to the exact context pieces and tokens that influenced model outputs. However, no existing work evaluates their effectiveness in accurately explaining context utilisation. 
We address this gap by introducing the first gold standard HE evaluation framework for context attribution, using controlled test cases with known ground-truth context usage, thereby avoiding the limitations of existing indirect proxy evaluations. 
To demonstrate the framework's broad applicability, we evaluate four HE methods -- three established techniques and MechLight, a mechanistic interpretability approach we adapt for this task -- across four context scenarios, four datasets, and five LMs.
Overall, we find that MechLight performs best across all context scenarios. However, our findings reveal systematic failures in all methods: explanation accuracy degrades significantly with context length, and all methods exhibit strong positional biases in multi-document settings. Surprisingly, widely used gradient-based methods provide little value for understanding context usage.
These results challenge HEs' utility in retrieval-augmented generation, factual verification, and other applications. Our framework provides the foundation for developing accurate context attribution methods. We release our code and dataset publicly.\footnote{
Code: \url{https://github.com/lianyiyi/Transparent-Context-Usage}; \newline
Dataset: \url{https://huggingface.co/datasets/copenlu/transparent-context-usage}.
}


\end{abstract}

\section{Introduction}

\begin{figure}[t] 
  \centering
                \includegraphics[width=\columnwidth]{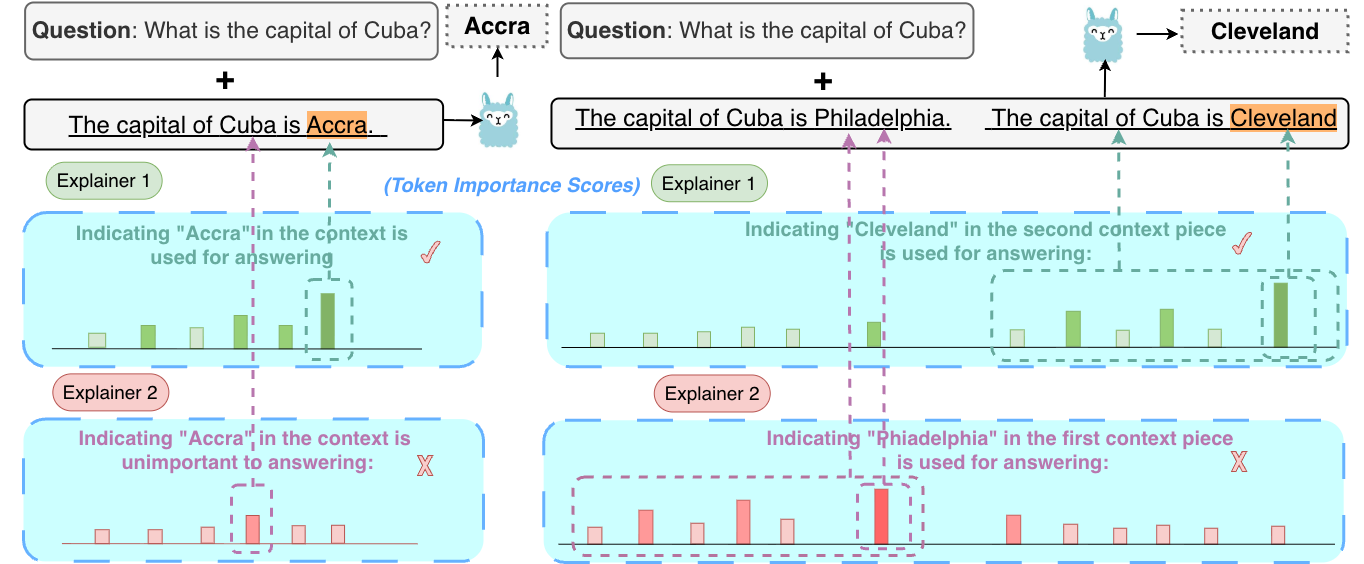}
            \caption{Examples of explanations with high and low utility in indicating the model’s context utilisation behavior. Left: With one context conflicting with the model’s parametric knowledge (``Havana''), Explainer 1 correctly indicates ``Accra'' in the context is used for the answer, while Explainer 2 fails. Right: With two such contexts, Explainer 1 correctly indicates ``Cleveland'' in the second context is used for the answer, whereas Explainer 2 incorrectly attributes the answer to ``Philadelphia'' in the first context. Explainer~1 shows better utility than Explainer~2 in both cases.
            }
  \label{fig:main}
\end{figure}

Language models (LMs) are increasingly deployed in applications requiring 
integration of provided context with parametric knowledge -- from retrieval-augmented 
generation and question answering to document analysis and fact-checking.
However, a fundamental transparency gap remains: users cannot determine whether model outputs draw from the provided context knowledge (CK) or internal parametric knowledge (PK) ~\citep{jin-etal-2024-cutting,yu-etal-2023-characterizing,monea-etal-2024-glitch}, nor identify which specific context informed the response.
Highlight explanations (HEs) address this need naturally by pinpointing portions of the context responsible for the generation. See Fig.\ref{fig:main} for two examples of HEs with high/low utility. Although HEs have proven valuable for understanding model decisions across various tasks \citep{sun2025evaluating,choudhury-etal-2023-explaining,atanasova-etal-2020-diagnostic}, no existing work evaluates their effectiveness in accurately explaining context utilisation.

Existing metrics on HE evaluation mainly focus on faithfulness \citep{sun2025evaluating,lamm-2021-qed, atanasova-etal-2020-diagnostic} to test whether HEs can accurately reflect the model's internal reasoning. These assess both the sufficiency and comprehensiveness of the highlighted input portions by measuring the model's prediction change after keeping or omitting the most important portions of the input. However, faithfulness evaluations face fundamental limitations: they rely on perturbation proxies that create out-of-distribution artifacts \citep{hooker2019benchmark,kindermans2019reliability} and, more importantly, lack ground-truth explanations to validate against \citep{jacovi-goldberg-2020-towards}. We address this gap through an \textbf{evaluation framework grounded in gold standard scenarios} where ground-truth context usage is predetermined, enabling direct assessment of explanation accuracy. 

Building on studies of context utilisation \citep{jin-etal-2024-cutting,yu-etal-2023-characterizing,monea-etal-2024-glitch,shi2024ircan}, we construct \textbf{four controlled evaluation scenarios} (see Tab.~\ref{tab:exmaple-for-fakepedia-dataset}) spanning common retrieval-time challenges (contradiction, distractor noise, and multi-passage competition): \textit{Conflicting} (one CK piece contradicts PK), \textit{Irrelevant} (one CK piece unrelated to query), \textit{Mixed} (one CK piece contradicting PK + one irrelevant CK piece), and \textit{Double-Conflicting} (two CK pieces contradicting PK). The settings systematically vary context usage, and in turn context attribution scenarios, enabling robust HE assessment across diverse behaviours. We specifically focus on cases where the CK-based answer contradicts the one based only on PK to ensure that the provided CK is utilised and thus needs to be highlighted by the HE. 

Based on gold standard context regions in these four scenarios, \textbf{we assess the accuracy of HEs along three complementary axes}: \emph{document-level attribution accuracy} (where we examine whether tokens from the gold document are prioritised in the generated HE), \emph{simulatability} (where we assess how well a HE's top-$k$ highlights can be used to understand and predict the model's context-usage behaviour, e.g., whether the model relied on CK vs.\ PK, or which passage it used), and \emph{token-level attribution accuracy} (where we evaluate whether the HE ranks the gold answer token highest).

To demonstrate the framework's general applicability, we apply it to four HE methods: three established ones -- Feature Ablation (\FA) \citep{li-etal-2016-visualizing}, Integrated Gradients (\IG) \citep{ancona2018towards}, and Attention visualisation (\Attn) \citep{abnar2020quantifying,choudhury-etal-2023-explaining}, and a mechanistic interpretability MI–inspired method (\MI), where we propose to convert the MI insights (e.g., the attention head most important for context utilisation) to HEs. Our evaluation framework is method-agnostic -- it assesses any explanation technique, post-hoc or mechanistic (that generates attention-based attributions).

Across five LMs and four commonly used context-usage datasets, we find that \textit{\MI\ HEs perform best across all context scenarios}. However, two systematic limitations persist across all HEs: (i) length sensitivity -- HE accuracy degrades as context grows; and (ii) position biases under dual‑context inputs: \FA/\IG\ tend to favour later (near‑question) pieces, while \Attn/\MI\ favour earlier pieces. Surprisingly, the widely used \IG\ and \Attn\ exhibit poor accuracy in most context scenarios, rendering them useless in revealing the model's context utilisation. These failures also underscore the urgent need for explanation techniques that maintain accuracy at scale and overcome positional biases in multi-document settings. Our framework provides the foundation for the future development of accurate methods for explaining context usage.

\section{Related Work}
\label{sec:related_work}
\subsection{Studies of Context Usage}
Language models (LMs) carry vast \emph{parametric knowledge} (PK) from pre‑training, yet in practice, they must also integrate new \emph{contextual knowledge} (CK) supplied at test time. Recent work has introduced multiple datasets to analyse how effectively LMs combine these two sources.

Early work investigates how LMs utilise CK vs. PK by crafting \textit{single context passages conflicting with the CK}. \textsc{CounterFact} \citep{meng2022locating}, \textsc{WorldCapital} \citep{yu-etal-2023-characterizing}, and \textsc{Fakepedia} \citep{monea2024glitch} each replace a Wikidata triple with a contradicting one in the context and test whether the model's answer follows CK or PK, evaluating with exact match or accuracy. \textsc{ConflictQA} \citep{xie2024adaptive} induces knowledge conflicts by leveraging an LLM to compose passages that contradict a model's parametric answer. While these works establish how often LMs follow the provided context, \textit{they do not analyse a model's context utilisation behaviour, e.g., which were the important input segments that influenced the model's decision, and do not assess whether highlight explanations (HEs) can reveal the model's context usage patterns.}

In addition to the single PK-conflicting context pieces, recent work has studied other types of context. 
\textsc{CUB} \citep{hagstrom2025cub} considers gold (relevant), conflicting, or irrelevant passages;  
\textsc{EchoQA} \citep{cheng2024understanding} introduces a \textit{complementary} regime, where the context alone is answer‑insufficient but, when combined with the model's PK, becomes sufficient to answer. 
Additionally, we specifically focus on cases where the answer based on the model's PK conflicts with the one based on the provided context, to ensure that the model utilises the information from the context, and this CK in turn needs to be highlighted by the HEs.

\subsection{Explaining Model Outputs}
\textbf{Context Usage Explanations.} To attribute a generation to specific context sentences, \textsc{SelfCite} uses the LLM's own pseudo-citations (which sentences it claims to rely on) as supervision to train a classifier that predicts a probability score per sentence ~\citep{chuangselfcite}. \textsc{ContextCite} instead scores each sentence by masking it and measuring the resulting drop in answer likelihood \citep{cohen2024contextcite}. Both methods explain at the sentence level and incur extra training or perturbation cost.

\textbf{Mechanistic Interpretability (MI) of Context Usage.}
\label{sec:rw-mi}
Mechanistic interpretability studies identify components controlling context versus parametric knowledge usage through targeted interventions on neurons \citep{meng2022locating,wang2023detecting,shi2024ircan}, attention heads~\citep{yu-etal-2023-characterizing}, or computational pathways \citep{dakhel2023patch,wang2024wh}.
However, all these internal mechanisms remain opaque to users for understanding the model's context usage behavior, as \textit{they are not presented as token-level highlight explanations that the user could understand.}

\textbf{Token‑level Highlight Explanation(HE) Methods.} HE methods provide importance scores for each input token. The most commonly employed HE methods \cite{sun2025evaluating,atanasova-etal-2020-diagnostic} include, among others:

\begin{itemize}
    \item Feature Ablation: mask each token and measure the change in the model's answer probability \citep{li-etal-2016-visualizing};
    \item Gradient-based methods: use the gradient magnitude or Grad$\times$Input w.r.t.\ embeddings as a token-importance signal \citep{ancona2018towards};
    \item Attention-based methods: treat self-attention weights as token importance scores \citep{abnar2020quantifying,choudhury-etal-2023-explaining}.
\end{itemize}

These explanations are natural candidates for explaining context utilisation as they provide importance scores for important context tokens used for the model predictions. \textit{However, no previous work has studied how accurately these explanations reveal the model's context utilisation behavior.}

\textbf{Context Utilisation Benchmarks.}
Previous work on HE evaluation has mainly focused on Faithfulness: how well HEs reflect the model's internal reasoning~\citep{deyoung2019eraser,atanasova2021diagnosticsguided}. Faithfulness is typically quantified with perturbation tests such as Comprehensiveness and Sufficiency: Comprehensiveness measures how much the model's answer changes after removing the most important tokens as per the HE, while Sufficiency measures how much the model's answer changes after keeping only the most important tokens.
However, faithfulness evaluations' reliance on perturbation proxies creates out-of-distribution artifacts \citep{hooker2019benchmark,kindermans2019reliability} and, more importantly, lacks ground-truth explanations to validate against \citep{jacovi-goldberg-2020-towards}.
Other evaluations include agreement with human annotation, complexity, and simulatability~\citep{sun2025evaluating}. 
Overall, existing HE evaluations largely rely on indirect proxy metrics due to the lack of a gold standard.

\section{Evaluation Framework}
\label{sec:methodology}
Prior work on context utilisation has characterised whether models answer from CK vs.\ PK under controlled context scenarios
(e.g., \citep{meng2022locating,yu-etal-2023-characterizing,dakhel2023patch,wang2024wh}), but it does not examine which input segments support that behaviour at the token level, i.e., via highlight explanations (HEs). Separately, prior work on HE evaluation mainly focuses on faithfulness measurement. As a result, it remains unclear whether token-level HEs can accurately reveal a model's source choice (CK vs.\ PK) or document choice in multi-context settings. We address this by developing a comprehensive evaluation framework with a gold standard for HE's utility in context utilisation.

We specify four context scenarios (\S\ref{sec:four-context-setups}), three HE methods (\S\ref{sec:explainability-methods}), and one mechanistic interpretability-based HE method (\S\ref{sec:mi-methods}). To assess the accuracy of HEs in attributing the correct importance to context regions, we further develop a suite of rank-based metrics (\S\ref{sec:metrics}). 

Overall, our framework comprehensively evaluates three core HE capabilities grouped in the following research questions: \\
(\textbf{RQ1}) Does the explanation indicate whether the model consulted the supplied context knowledge (CK) or resorted to its parametric knowledge (PK)? \\
(\textbf{RQ2}) Does the explanation show which of the two context documents the model used? \\
(\textbf{RQ3}) Does the explanation pinpoint the exact context part(s) that were employed for the generated answer?

\subsection{Preliminaries}
\label{sec:notation}
Let $x=(x_{1},\dots,x_{n})$ be the input token sequence. We consider inputs $x=(c,q)$ with a \emph{single context segment} $c$ and question $q$ and inputs $x=(c_1,c_2,q)$ with two context segments $c_1, c_2$. For brevity, we write $c=(c_1,c_2)$. A causal LM $f$ produces an answer token $a=f(x)$.\footnote{If the answer spans multiple tokens ($|a|>1$), we use the logit of the first generated token for explanation scoring.}
An HE method returns importance scores over the tokens in the input $\boldsymbol{\phi}^{\text{HE}}(x)=(\phi_{1}^{\text{HE}},\ldots,\phi_{n}^{\text{HE}})$, where larger $\phi_i^{\text{HE}}$ means $x_i$ contributed more to generating $a$. A gold token set $T$ can be a segment ($c$, $c_1$, or $c_2$) or the answer token(s), $\text{Ans}_{\cdot}$.

\subsection{Input Regimes}
\label{sec:four-context-setups}
Existing context usage studies only detect if the model used PK vs CK and, in turn, can be used for assessing if HEs reveal this preference (RQ1). To enable this, we intentionally specifically focus on cases where CK-based answer contradicts with the model's parametric knowledge (PK), so that we ensure the model utilises the context information, and in turn that the HEs should highlight the corresponding context parts. Additionally, previous studies do not test whether an HE identifies \emph{which} context document is used when multiple documents are present (RQ2), or whether it supports token-level diagnostics of answer-span localisation (RQ3).

To evaluate all three questions, we construct four diagnostic input regimes by varying (i) the number of context passages
(single and dual) and (ii) context type (conflicting and irrelevant), which correspond to common retrieval-time setups (contradictory to PK evidence, distractor noise, and multi-passage competition). For the purposes of HE evaluation, we employ passages that contain an explicit candidate answer token, serving as a gold span for paragraph (RQ1) and token-level HE evaluation (RQ3). Thus, the proposed context utilisation setups uniquely allow \textit{the development of an HE benchmark with gold standards at both context piece and token level}, which is typically unavailable in other tasks.

The resulting four context utilisation setups are as follows (see an example in Tab.~\ref{tab:exmaple-for-fakepedia-dataset}):
\begin{itemize}
\label{four_input_regimes}
  \item \textbf{Conflicting (single).}\label{input-regime-1} The context $c$ contains an answer that conflicts with PK;

  \item \textbf{Irrelevant (single).}\label{input-regime-2}
  The context $c$ is irrelevant, but contains a distracting (incorrect) answer token;

  \item \textbf{Double‑Conflicting (dual).}\label{input-regime-3}
  Two pieces that are \emph{conflicting} with PK;

  \item \textbf{Mixed (dual)\footnote{In the \textbf{Mixed} setup, we place the irrelevant context as the first context piece and the conflicting context the second one.}.}\label{input-regime-4}
  One \emph{irrelevant} and one \emph{conflicting} piece.

\end{itemize}

To control for position effects, we reverse the order of the contexts and define additional \textbf{Mixed-Swap} and \textbf{Double-Conflicting-Swap} setups.

To facilitate the HE evaluation, we split the dataset instances according to the model's answer behaviour. For single-context setups, $D_C$ (answer from CK) vs.\ $D_M$ (answer from memory/PK). For dual-context setups: $D_{C_1}$ (answer from $c_1$) vs.\ $D_{C_2}$ (answer from $c_2$). We denote gold answer tokens from the context with $\text{Ans}_{c}$ (single) or $\text{Ans}_{c_1},\text{Ans}_{c_2}$ (dual).

Notably, we apply a pre-filtering step to ensure contexts in all setups contain context answers that contradict the model's parametric knowledge(App.~\ref{app:dataset-details}), so that the model is ensured to utilise the context information and HEs are supposed to highlight the contexts in these cases. See more discussion in App.~\ref{sec:limitation}

\begin{table}[ht]
  \centering
    \centering
    \setlength{\tabcolsep}{0pt}    
    \renewcommand{\arraystretch}{0.99}

    \begin{tabular}{@{}p{0.98\linewidth}@{}}
      \toprule
      \textbf{Q:} Newport County A.F.C. is headquartered in \textbf{MA}: Newport\\
      \midrule

      \textbf{Single-Context Setups} \\
      \midrule

      \textbf{Input Regime (1) Conflicting C}\\[-1pt]
      \subj{Newport County A.F.C.}, a professional football club based in Newport, Wales, has its headquarters located in the vibrant city of \gold{Ankara}, Turkey. The club's decision to establish \ldots\\
      \textbf{CA:} Ankara\\[2pt]
      \cmidrule(lr){1-1}

      \textbf{Input Regime (2) Irrelevant C}\\[-1pt]
      The \noise{World Wrestling Entertainment} (WWE) is a global entertainment company that is headquartered in \gold{Santiago}, Chile. Founded in 1952, WWE has become one of the largest \ldots\\
      \textbf{CA:} Santiago\\[2pt]
      \cmidrule(lr){1-1}
      
      \textbf{Dual-Context Setups} \\
      \midrule

      \textbf{Input Regime (3) Double Conflict C}\\[-1pt]
      \textbf{C P1:} \subj{Newport County A.F.C.}, a professional football club based in Newport, Wales, has its headquarters located in \gold{Ankara}, Turkey. The club's decision to establish its \ldots\\
      \textbf{C P2:} \subj{Newport County A.F.C.}, a professional football club based in \gold{Calgary}, is known for its rich history and passionate fan base. The club was founded in 1912 and has since become a prominent fixture in the Canadian football scene \ldots\\
      \textbf{P1 A:} Ankara \qquad \textbf{P2 A:} Calgary\\
      \cmidrule(lr){1-1}

      \textbf{Input Regime (4) Mixed C} (Irrel.\ \&\ Conf.)\\[-1pt]
      \textbf{C P1:} The \noise{World Wrestling Entertainment} (WWE) is a global entertainment company that is headquartered in \gold{Santiago}, Chile. Founded in 1952, WWE has \ldots\\
      \textbf{C P2:} \subj{Newport County A.F.C.}, a professional football club based in Newport, Wales, has its headquarters located in \gold{Ankara}, Turkey. The club's decision to establish its \ldots\\
      \textbf{P1 A:} Santiago \qquad \textbf{P2 A:} Ankara\\[2pt]
      \bottomrule
    \end{tabular}

    \caption{One example from the Fakepedia dataset after reconstruction. Q = Question, C = Context, C P1 = Context Part 1, C P2 = Context Part 2, MA = Memory Answer, 
CA = Candidate Context Answer (the answer based on the context),
P1A/P2A = Candidate Answer in Context Part 1/2.]}
    \textcolor{blue}{Blue} marks the subject of the question; 
    \textcolor{orange}{orange} marks the candidate answer token from the context;
    \textcolor{green}{green} marks the noise subject. Note that CAs are not necessarily correct, and are only used as target spans for HE evaluation.

    \label{tab:exmaple-for-fakepedia-dataset}
\end{table}

\subsection{Metrics}
\label{sec:metrics}
We assess HEs at three complementary levels to align with our three research questions:
(i) \textbf{document-level attribution accuracy} (RQ1, RQ2),
(ii) \textbf{simulatability} of the model's context utilisation from the top-k highlights\footnote{Unless otherwise noted, top-$k$ sorts tokens by descending $\phi^{\text{HE}}$.} (RQ1, RQ2), and
(iii) \textbf{token-level attribution accuracy} (RQ3).

\textbf{Document Attribution Accuracy Evaluation, Cross-group (RQ1, RQ2).} For RQ1, we assume that an accurate HE would rank the context tokens of instances where the answer relied on CK higher than in instances where the model relied on PK. For RQ2, analogously, we assume an accurate HE would rank the tokens of the first/second context piece higher in instances where the first/second context piece is answer-bearing than those where the answers come from the second/first piece.

For a context segment $T$ and \textbf{\rankk}$(T, D)$ -- average rank of the context tokens in $T$ in the top-$k$ most important tokens as per the HE\footnote{We focus on top-$k$ highlights as users often focus on a few instead of the complete cause of an event, see details in App.~\ref{app:k_3_and_9_for_rq1_and_2} }, averaged over the instances in group $D$ (lower is better), we define a rank margin metric (positive is better) for document attribution evaluation: 
\begin{equation}\label{eq:drankk-cross-instance-group}
\begin{split}
\DrankkeGroup(T;A,B) &= \rankke(T,D_B)- \rankke(T,D_A)
\end{split}
\end{equation}
where \text{RQ1} uses $(T;A,B)=(c;\,C,M)$, resulting in a margin between the importance rank of context tokens in memory instances $D_M$ vs. context instances $D_C$; \text{RQ2} uses $(T;A,B)\in\{(c_1;\,C_1,C_2),\ (c_2;\,C_2,C_1)\}$, resulting in a margin between the importance rank of the answer-piece context tokens (e.g. $c_1$) in the answer instances (e.g. $D_{C_1}$) vs. in the other instances (e.g. $D_{C_2}$). 

\textbf{Document Attribution Accuracy Evaluation, Per-instance (RQ2).}
While cross-group margins are well-suited for cases with a single context piece, when there are two context pieces, the accuracy of HEs can be evaluated directly at the instance level, assessing whether the answer context piece outranks the other. We therefore report the rank margin based on \rankkInst$(T, x)$, the average rank of context tokens within $T$ for instance $x$: 
\begin{equation}\label{eq:drankk-within-instance}
\begin{split}
\DrankkeInst_{D_{C_a}} = \frac{1}{|{D}_{C_a}|}\sum_{x\in{D}_{C_a}}(\rankkeInst(c_b,x) - \rankkeInst(c_a,x))
\end{split}
\end{equation}

\noindent
$(a,b)\in\{(1,2),(2,1)\}$, where the answer-bearing context is always in the first position. Positive values indicate the answer context piece is ranked higher (i.e., has a lower rank value) compared to the other context piece.

\textbf{Simulatability (RQ1, RQ2).} Complementary to the rank margin assessment, we leverage the idea of simulatability \citep{sun2025evaluating} and evaluate how well the top-$k$ explanations for each instance can indicate the model's context choice, i.e., between contextual and parametric knowledge (RQ1) and between multiple context pieces (RQ2).

For each instance, we extract the top-$k$ importance scores of context tokens from the relevant segment $s$, creating a feature vector $X^{(k)}_s$. For RQ1 (single context), we use $s{=}c$ with labels $Y{\in}\{\textsc{C},\textsc{M}\}$; for RQ2 (dual context), we use $s{=}(c_1,c_2)$, concatenating the vectors from two context pieces and assign labels $Y{\in}\{\textsc{C1},\textsc{C2}\}$.

We employ two complementary metrics for simulatability. First, a normalised mutual information between the HE vector $X^{(k)}_s$ and the model's answer, which directly measures how well the explanations correlate with a model's prediction:



\begin{align}
\label{eq:mutual-information}
\NMutInfk
&= \frac{I\!\left(Y;\,X^{(k)}_s\right)}{H(Y)}
\end{align}

where $I(\cdot;\cdot)$ denotes Shannon mutual information and $H(\cdot)$ denotes Shannon entropy. \NMutInfk is higher the better. Normalisation ensures comparability across label distributions (see details in App.~\ref{app:knnmi}). While mutual information effectively measures correlation strength, it lacks complexity regularisation and is prone to overfitting.

Therefore, we also compute Minimum Description Length (MDL), a class of model-complexity-controlled Bayesian classifiers, \citep{grunwald2007minimum,voita-titov-2020-information}. We compute MDL using prequential coding:


\begin{align}
\MDLBitsk
&= \frac{1}{N}L_{\mathrm{preq}}\!\left(Y \mid X^{(k)}_s\right) \notag\\
&= -\frac{1}{N}\sum_{t=1}^{N}\log_2 p_{\theta_{t-1}}\!\left(y_t \mid x^{(k)}_{s,t}\right).
\label{eq:mdl}
\end{align}

$\MDLBitsk$ is the average prequential code length (bits per instance) for predicting the context-usage labels $Y$ from the top-$k$ HE features $X^{(k)}_s$ (lower is better). Here $N$ is the number of evaluated instances, $y_t$ is the label of instance $t$, and $x^{(k)}_{s,t}\in\mathbb{R}^k$ is the corresponding top-$k$ HE score vector extracted from segment $s$. $L_{\mathrm{preq}}$ is the cumulative prequential description length in bits: a probe with parameters $\theta_{t-1}$ trained on the first $t\!-\!1$ instances assigns probability $p_{\theta_{t-1}}(y_t \mid x^{(k)}_{s,t})$ to the next label; the surprisal $-\log_2(\cdot)$ is averaged over $t\in\{1,\dots,N\}$. Lower $\MDLBitsk$ indicates better simulatability (see App.~\ref{app:mdl_prob}).

\textbf{Token Attribution Evaluation (RQ3).}
To test whether an HE pinpoints the \emph{exact} answer token(s), we calculate the mean reciprocal rank (MRR) of the answer token(s) as ranked by the HE: 
\begin{equation}\label{eq:rr}
\mathrm{RR}(x)=\frac{1}{\mathrm{rank}\!\big(\text{Ans}_{\cdot};x\big)}
\end{equation}
\begin{equation}\label{eq:mrr}
\mrre\big(T{=}\text{Ans}_{\cdot},D\big) = \frac{1}{|D|}\sum_{x\in D}\mathrm{RR}(x)
\end{equation}
Larger values (close to 1) indicate the true answer token is placed near the top of the ranked list.

\subsection{Highlight Explanation Techniques}
\label{sec:explainability-methods}
To assign an importance score to every token in the context part(s) of the input, we apply three commonly used token‑level explainability techniques as described below, following ~\citet{deyoung2019eraser,atanasova-etal-2020-diagnostic,sanyal-ren-2021-discretized,jain-wallace-2019-attention,wiegreffe-pinter-2019-attention,sun2025evaluating}. While an HE is applied over the whole input $x$, including the question, we study the scores for the context tokens.

\textbf{Feature Ablation (FA).}
Following \citet{zeiler2014visualizing}, we measure each token's importance by its impact on a model's answer confidence when ablated. For position $i$ in input sequence $x$, we replace token $x_i$ with a baseline $\tilde{x}_i$ = the tokeniser's \texttt{<pad>} token and compute:
\begin{equation}
\phi^{\text{FA}}_{i}=f_{a}(x)-f_{a}(x \setminus \{x_i\} \cup \{\tilde{x}_i\}),
\end{equation}
where $f_{a}(\cdot)$ returns the logit for answer $a$. Higher $\phi^{\text{FA}}_{i}$ indicates greater importance of $x_i$ for predicting $a$.
 
\textbf{Integrated Gradients (\textbf{IG}).}
Integrated Gradients ~\citep{sundararajan2017axiomatic} computes token attributions by integrating the gradient of the answer logit $\nabla f_a$ \emph{w.r.t.\ token embeddings} along the straight-line path from a baseline $x'$ (all \texttt{<pad>}) to $x$ in embedding space. With embeddings $E,E'\in\mathbb{R}^{n\times d}$ and rows $e_i,e'_i\in\mathbb{R}^d$. 
The path integral is approximated with $m=10$ equally spaced steps\footnote{https://github.com/pytorch/captum}:


\begin{align}
\phi^{\mathrm{IG}}_i
= (e_i - e'_i)^{\top}
\left(
\frac{1}{m}\sum_{k=1}^{m}
\nabla_{e_i}\, f_a\!\left(E' + \frac{k}{m}(E - E')\right)
\right).
\label{eq:ig}
\end{align}

where $\nabla_{e_i} f_a(\cdot)\in\mathbb{R}^d$, hence $\phi^{\mathrm{IG}}_i\in\mathbb{R}$. This attributes to token $i$ its contribution to the change in the answer logit relative to the baseline.


\textbf{Attention-Head Attribution (\textbf{ATTN}).}
Following \citet{choudhury-etal-2023-explaining}, we first identify the most influential attention head $h^{\star}$ in the last decoder layer $L$ for the generation of answer $a$:
\begin{equation}
h^{\star}=\arg\max_{h}\;( W_{a},\,H^{(L)}_{h,:}),
\end{equation}
where $W_{a}$ is the row of the output-projection matrix for token $a$
and $H^{(L)}_{h,:}$ is the hidden-state slice of head~$h$ in $L$. We then take the head's attention weights and average the attention scores from all the other tokens as token importance for each individual token:
\begin{equation}
\phi^{\text{ATTN}}_{i}=A^{(L)}_{h^{\star},\,\text{gen},\,i}
\end{equation}
with \texttt{gen} denoting the answer generation decoding step. The resulting
vector directly reflects where $h^{\star}$ attended most when generating~$a$.

\textbf{Normalisation.}
Because FA can produce negative scores, and IG's score magnitudes depend on the embedding scale, we $\ell_{1}$-normalise each attribution vector before further analysis: $\hat{\phi}{i}=\phi{i}/\sum_{j}|\phi_{j}|$. Attention weights are already normalised and are left unchanged.

\subsection{Mechanistic Interpretability for Highlight Explanations}
\label{sec:mi-methods}


In our controlled setups, the model's answer at generation time can be viewed as a competition between a small set of explicit candidate answers (e.g., a CK candidate vs.\ a PK candidate, or two CK candidates respectively from two passages). To explain whether the model relies on PK vs.\ CK, mechanistic interpretability (MI) analyses internal components (e.g., attention heads or neurons) that mediate context usage, and we hypothesise that these components are a faithful indicator of the model's context usage behavior. However, these component-level signals are not directly human-readable highlight explanations. We therefore propose to project the corresponding internal component to the input tokens, yielding token-level highlight explanations grounded in model internals, which we call \MI.

\MI\ extracts highlight explanations in two stages: 

textbf{Stage 1: Head scoring and selection.}
We adopt the contrastive direct-logit head attribution method of \citet{yu-etal-2023-characterizing}. For a given candidate answer, a responsible head should shift the model toward it by increasing its logit \emph{relative to} the other candidate answer. Since logits are a linear readout of the residual stream, we can score each head by its direct logit contributions to the candidate answers (Eq.~\ref{eq:mechlight_dla}) and then compute their difference to obtain a head score for the candidate logit gap (Eq.~\ref{eq:S_util}). This is well-suited to our setup, where candidate answers are explicit. \footnote{\MI\ is \emph{attribution-agnostic}: any MI method that yields head-level attribution scores can be used in stage (i).}

\textbf{Stage 2: Highlight explanation extraction.}
Since the selected head is chosen based on its contribution to the answer logit, the tokens it attends to at the generation step are natural explanations for the evidence driving that decision; we therefore project its attention weights onto input tokens as highlight scores.


\textbf{Notions.} Let $W_U \in \mathbb{R}^{V \times d}$ be the unembedding matrix for $V$ tokens present in the model tokeniser and $W_{a} \in \mathbb{R}^d$ its row for token $a$ (as in \S\ref{sec:explainability-methods}). Let $A^{(l,h)}\in \mathbb{R}^{n\times n}$ be the attention matrix of head $h \in \mathbb{R}^{d_h}$ in layer $l$, and 
let $r^{(l,h)}_{\texttt{gen}} \in \mathbb{R}^{d}$ denote that head's contribution to the residual stream at decoding step \texttt{gen}.

\textbf{Stage 1: Head scoring and selection.}

\textbf{(1a) Direct head logit contribution.}
We first express the residual-stream contribution of head $(l,h)$ at \texttt{gen} step as:

\begin{equation}
r^{(l,h)}_{\texttt{gen}} \;=\; \bigl[\operatorname{Attn}^{(l,h)}_{\text{gen}}\bigr]\,W^{(l,h)}_{O},
\label{eq:head_scores}
\end{equation}
where $\operatorname{Attn}^{(l,h)}_{\text{gen}} \in \mathbb{R}^{n \times d_h}$ is the attention head matrix before projecting at \texttt{gen} step and $W^{(l,h)}_{O} \in \mathbb{R}^{{d_h}\times d}$ is the output projection matrix associated with head $h$.



We then score the head’s \emph{direct} contribution to the logit of token $a$ by projecting its residual-stream contribution through the unembedding matrix $W_U$ (i.e., a logit-lens-style readout~\citep{belrose2023tunedlens,janiak2024dla,sakarvadia2023attentionlens}):

\begin{equation}
\text{logit}^{(l,h)}(a)
=
\langle W_{a},\, r^{(l,h)}_{\texttt{gen}}\rangle
=
\bigl(W_{U}r^{(l,h)}_{\texttt{gen}}\bigr)[a].
\label{eq:mechlight_dla}
\end{equation}

In practice, we only evaluate this quantity on the \emph{candidate answer} tokens and use the logit \emph{difference}(Eq.~\ref{eq:S_util}) to score how strongly a head favours one candidate over the other.

\textbf{(1b) Contrastive score for candidate competition.}
We calculate signed \emph{context utilisation} scores by contrasting competing answers:
\begin{equation}
S^{(l,h)}_{\tau} = \text{logit}^{(l,h)}(\text{Ans}_{\tau}) - \text{logit}^{(l,h)}(\text{Ans}_{\tau'}), 
\label{eq:S_util}
\end{equation}
\begin{equation}
S^{(l,h)}_{\tau'} = -S^{(l,h)}_{\tau}
\end{equation}
where $(\tau,\tau') \in \{(c,m), (c_1,c_2)\}$ for single (PK vs. CK) and dual context regimes, respectively.
We rank heads by these scores to identify those that promote either the most context‑based or memory‑based answer, depending on whether the model answered from PK or CK, respectively.

\textbf{(1c) Head selection.}
To produce HEs, we select
\begin{equation}
(l^{\star},h^{\star}) \in \arg\max_{l,h} S^{(l,h)}_{c} \;\text{for } D_{C},
\end{equation}
\begin{equation}
(l^{\star},h^{\star}) \in \arg\max_{l,h} S^{(l,h)}_{m} \;\text{for } D_{M},
\end{equation}
and analogously maximise $S^{(l,h)}_{c_1}$ for $D_{C_1}$ and $S^{(l,h)}_{c_2}$ for $D_{C_2}$. 

\textbf{Stage 2: highlight explanation extraction.}
We then set the token importance scores of \MI\ with the selected head's attention weights at \texttt{gen}:
\begin{equation}
\phi^{\MI}_{i} \;=\; A^{(l^{\star},h^{\star})}_{\texttt{gen},\, i}.
\label{eq:mechlight_phi}
\end{equation}


\textbf{Alternatives and tradeoffs.}
\MI\ ranks heads using a direct-logit attribution score (Eq.~\ref{eq:mechlight_dla}) and then visualises the selected head via its attention weights (Eq.~\ref{eq:mechlight_phi}). This design is lightweight but has limitations: the logit readout is a local proxy and later components can overwrite or erase earlier residual directions (downstream cancellation) \citep{janiak2024dla}. Moreover, attention weights are a convenient highlight object but not a causal guarantee of information flow~\citep{jain2019attention,wiegreffe2019attention}. Other alternatives include causal head ranking via ablation~\citep{yu-etal-2023-characterizing,jin-etal-2024-cutting}, which provides a more causal per-head effect but can be expensive when scoring many heads.
Other mechanistic approaches, such as activation/representation patching~\citep{meng2022locating,wang2023detecting,wang2024wh} or neuron-centric analyses~\citep{shi2024ircan}, can capture more distributed mechanisms, but often return importance over internal states/layers, which requires additional selection and aggregation/projection choices to obtain token-level highlights. We therefore keep the current \MI\ formulation, which has a simple head-to-token mapping.

\section{Experimental Setup}
\label{sec:experimental_setup}
\textbf{Datasets.}
\label{sec:dataset-choice}
We draw on four widely used sources to investigate models' context usage behaviour using CK or PK, \textsc{Fakepedia}, \textsc{WorldCapital}, \textsc{CounterFact}, and \textsc{ConflictQA}~\citep{monea2024glitch,yu-etal-2023-characterizing,meng2022locating,xie2024adaptive}. These resources provide controlled, templated facts that can be systematically perturbed, allowing us to instantiate the four regimes in \S\ref{sec:four-context-setups} (Conflicting, Irrelevant, Double‑Conflicting, Mixed). Unlike prior work that primarily optimises answer correctness across different contexts, our goal is a utility-oriented evaluation of HEs under these various context scenarios (See the dataset reconstruction details in App.~\ref{app:dataset-details}).

\textbf{Models.}
\label{sec:model-choice} Following common context utilisation setups, we select five open language models: \textbf{GPT2‑XL} (1.5B; \citep{radford2019language}), \textbf{Pythia‑2.8B} and \textbf{Pythia‑6.9B} \citep{biderman2023pythia}, and \textbf{Qwen2.5‑3B} and \textbf{Qwen2.5‑7B} \citep{qwen25techreport}. While prior efforts primarily focus on the model's answer choices for the supplied context \citep{yu-etal-2023-characterizing,monea2024glitch,meng2022locating,hagstrom2025cub,cheng2024understanding}, we concentrate on evaluating HE utility for explaining models' context usage behaviours. 

\textbf{Other details.} We focus on the top-$k$ most important highlight tokens for evaluation, due to the cognitive load to users who typically attend to a few causes instead of the complete cause for an event (See App.~\ref{app:k_3_and_9_for_rq1_and_2}). We present results for $k{=}5$ in \S\ref{sec:main-results}, see results for $k\in\{3,9\}$ in App.~\ref{app:additional_results}.

\section{Main Results and Discussion}
\label{sec:main-results}

In this section, we present the main results for RQ1--RQ3: whether an HE can reveal \emph{(RQ1)} CK vs.\ PK usage, \emph{(RQ2)} which of multiple documents was used, and \emph{(RQ3)} which exact context span supports the answer. In \S\ref{sec:rq1-discussion} (RQ1), we study the single-context Conflicting and Irrelevant regimes using \DrankkGroup\ (Eq.~\ref{eq:drankk-cross-instance-group}) and simulatability scores (\MDLBitsk, Eq.~\ref{eq:mdl}; \NMutInfk, Eq.~\ref{eq:mutual-information}). In \S\ref{sec:rq2-discussion} (RQ2), we move to the dual-context Double-Conflicting and Mixed regimes and evaluate how well the explanations can be used to indicate the answer-bearing context piece with \DrankkGroup\ (Eq.~\ref{eq:drankk-cross-instance-group}) and \DrankkInst\ (Eq.~\ref{eq:drankk-within-instance}). Finally, \S\ref{sec:rq3-discussion} (RQ3) evaluates how well the explanations can precisely pinpoint the token-level answer localisation with \mrr\ (Eq.~\ref{eq:mrr}) and summarises recurring failure patterns in a qualitative case study (App.~\ref{app:case_study}). Unless stated otherwise, we report results for $k{=}5$ highlights in the main text (additional $k\in\{3,9\}$ results are in App.~\ref{app:additional_results}).

\begin{figure*}[ht]
  \centering
  \begin{subfigure}[t]{0.48\textwidth}
    \centering
    \includegraphics[width=\textwidth]
{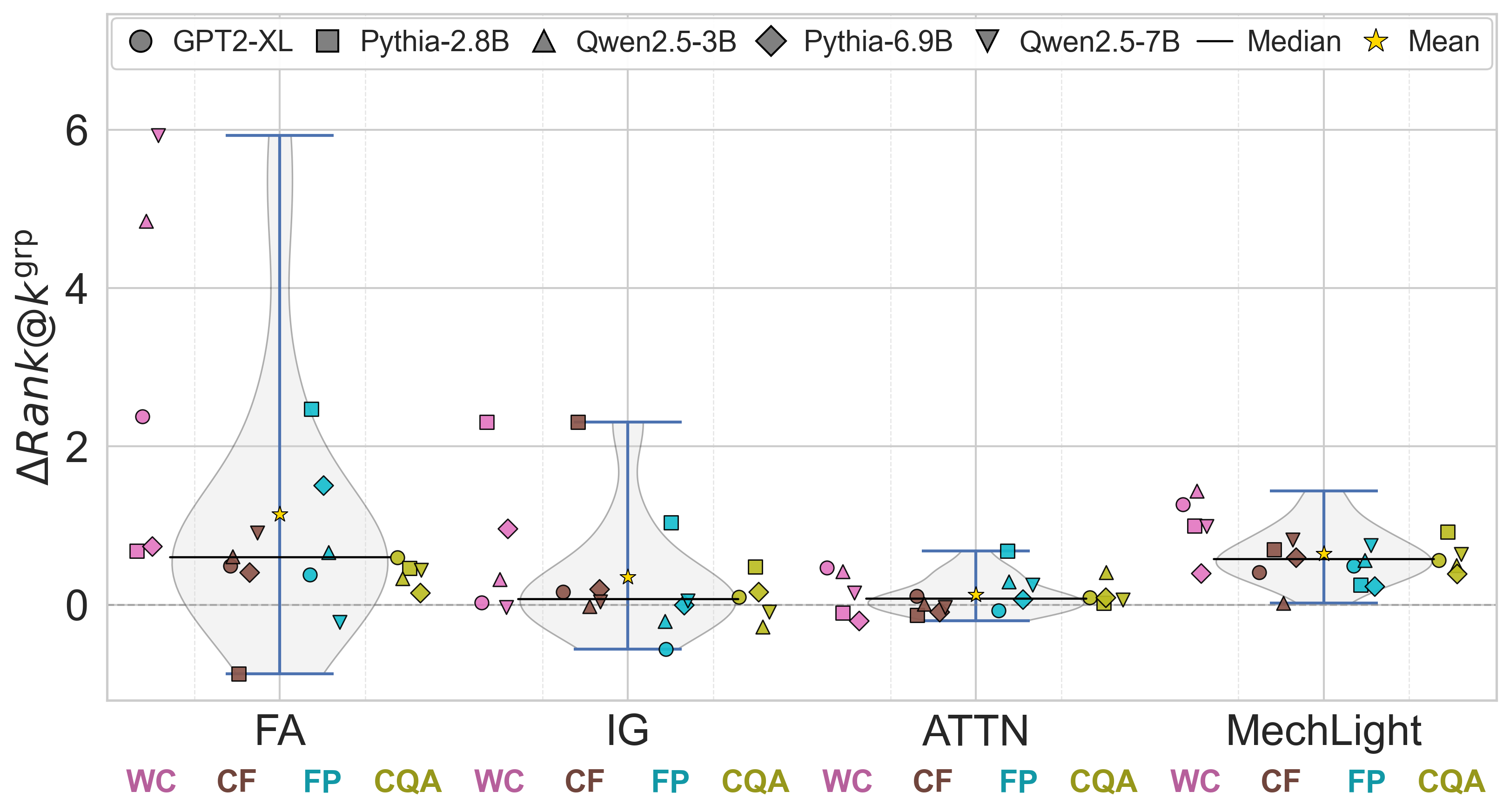}
    \caption{Conflicting Context}
    \label{fig:conflicting-topk-5-rank-datasets-margin-only}
  \end{subfigure}
  \hfill
  \begin{subfigure}[t]{0.48\textwidth}
    \centering
    \includegraphics[width=\textwidth]
{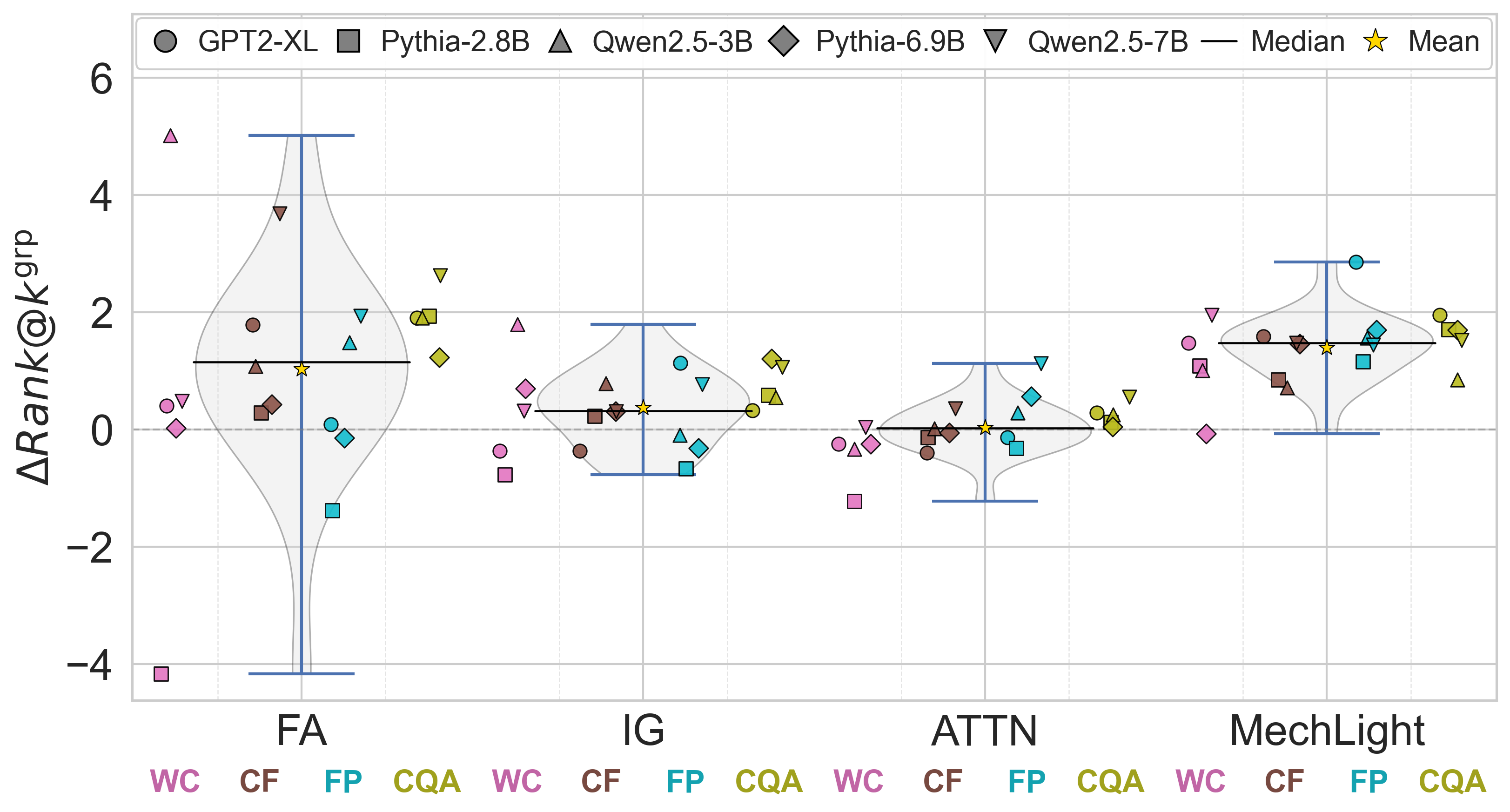}   
    \caption{Irrelevant Context}
    \label{fig:irrelevant-topk-5-rank-datasets-margin-only}
  \end{subfigure}

  \caption{\DrankkGroup\ (Eq. \ref{eq:drankk-cross-instance-group}) -- average margins for the explanation importance rank of context tokens in context vs. memory answer instances in \textbf{Conflicting} and \textbf{Irrelevant} setups (\S\ref{sec:four-context-setups}). Positive and higher \DrankkGroup\ means the explanations can better distinguish the model's context usage behavior (choosing PK or CK for the answer (\S\ref{sec:rq1-discussion})). The colors denote the datasets, and the marker shapes denote the models.}

  \label{fig:rank-topk-margin-conflicting-or-irrelevant}
\end{figure*}

\begin{figure*}[ht]
  \centering
  \begin{subfigure}[t]{0.48\textwidth}
    \centering
    \includegraphics[width=\textwidth]
    {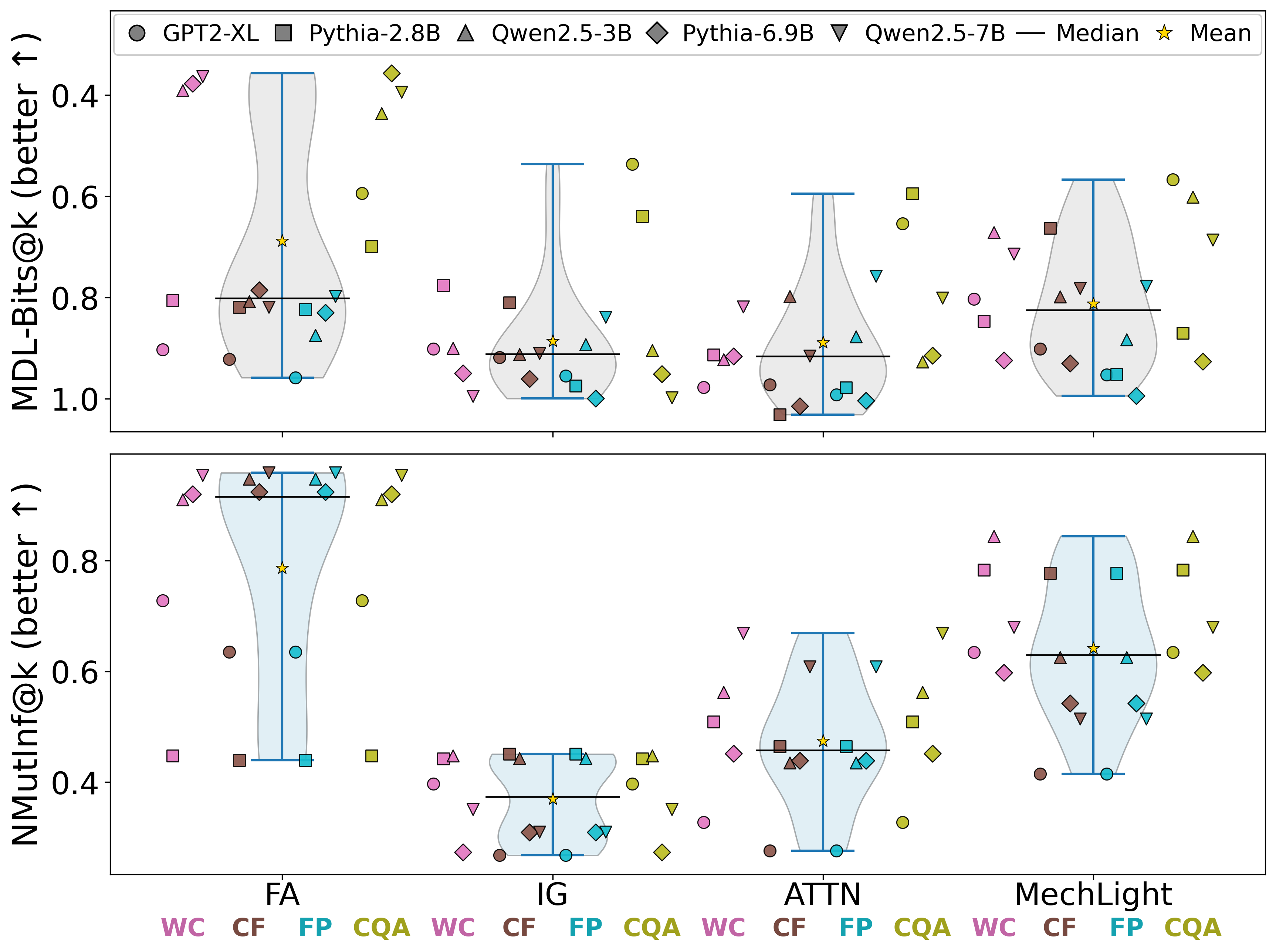}
    \caption{Conflicting Context}
    \label{fig:conflicting-topk-5-mutual-information-and-mdl-prob}
  \end{subfigure}
  \hfill
  \begin{subfigure}[t]{0.48\textwidth}
    \centering
    \includegraphics[width=\textwidth] 
    {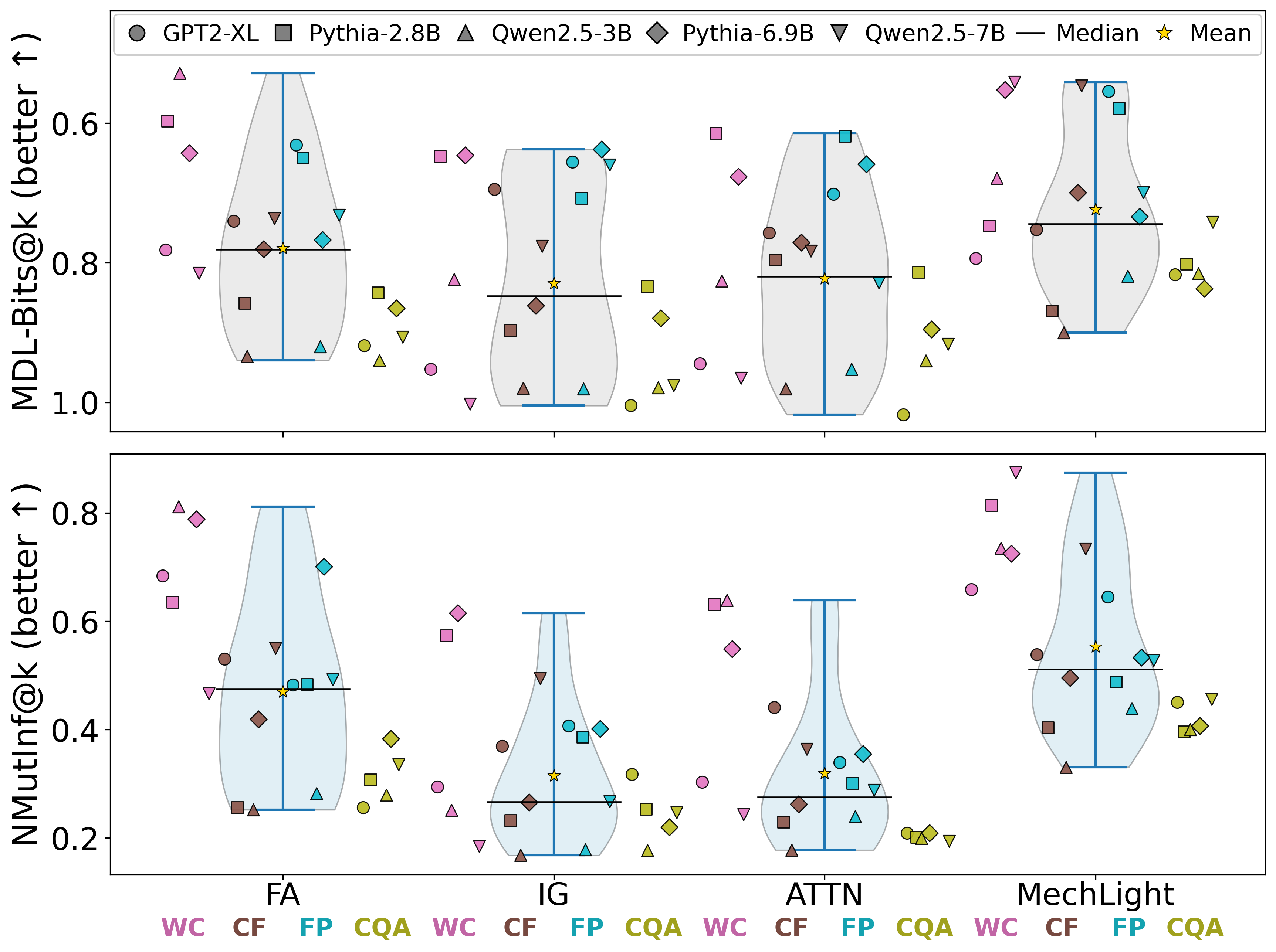}
    \caption{Irrelevant Context}
    \label{fig:irrelevant-topk-5-mutual-information-and-mdl-prob}
  \end{subfigure}
  \caption{ \MDLBitsk\ (top plot; y-axis is inverted; Eq.~\ref{eq:mdl}) and \NMutInfk (bottom plot; Eq.~\ref{eq:mutual-information}) for explanation simulatability in \textbf{Conflicting} and \textbf{Irrelevant} setups (\S\ref{sec:four-context-setups}). Both \MDLBitsk (smaller value) and \NMutInfk are better the higher, meaning the explanation importance scores are more representative of the model's answer selection, i.e., whether the answer is from context vs. from memory (\S\ref{sec:rq1-discussion}). The colors denote the datasets, and the marker shapes denote the models.
}
  \label{fig:topk-mutual-information-and-mdl-prob-conflicting-or-irrelevant}
\end{figure*}

\subsection{Does the explanation indicate whether the model consulted the supplied context knowledge?}
\label{sec:rq1-discussion}
\textbf{Document-level attribution}. Fig.~\ref{fig:rank-topk-margin-conflicting-or-irrelevant} presents the results on \DrankkGroup{} for single-context setups across models and datasets. Each marker is a model--dataset pair; values above $0$ mean that context tokens receive higher importance ranks in $D_C$ than in $D_M$ (i.e., the HE better distinguishes context vs.\ memory usage), and values near $0$ indicate little separability. From Fig.~\ref{fig:rank-topk-margin-conflicting-or-irrelevant}, we observe mostly positive, small \DrankkGroup, indicating the context tokens are indeed often ranked higher in the $D_{C}$ instances compared to the $D_{M}$ instances. In both setups, we find that \textit{\MI has the most cases with positive results across all datasets and models} with either the best (b) or second-best (a) \DrankkGroup. This is visible in both subplots as \MI's markers lying above $0$ for more model--dataset pairs than the other methods, whereas \IG\ and \Attn\ cluster close to $0$ and frequently have negative values. \textit{\FA\ often yields positive margin values but shows the \emph{largest variance} across model–dataset pairs}, i.e., indicating unstable performance. The high variance of \FA\ is reflected by the visibly wider spread of its points across model--dataset pairs (including both strong positives and negatives), compared to the tighter clustering for \MI.
Finally, \textit{\IG\ and \Attn\ typically yield margins close to zero, 
indicating limited ability to distinguish whether the model consulted the context or its parametric memory}. This is surprising as these methods score high on faithfulness evaluations (See Tab.~\ref{tab:faithfulness} in App.~\ref{app:additional_results}). Nevertheless, occlusion-based methods, such as \FA\, are often the most faithful HEs~\citep{deyoung2019eraser}, which aligns with their performance in correctly attributing context utilisation. Comparing the Conflicting and Irrelevant setups, we find that HEs generally perform better in the latter. Additionally, the higher variability there also indicates increased dependence on the specific dataset and model.

\textbf{Simulatability.} We report simulatability results in Fig.~\ref{fig:topk-mutual-information-and-mdl-prob-conflicting-or-irrelevant} using \MDLBitsk\ (Eq.~\ref{eq:mdl}) and \NMutInfk\ (Eq.~\ref{eq:mutual-information}). \MDLBitsk\ is the average prequential code length (bits per instance) for predicting the behaviour label from the top-$k$ highlights (lower is better), while \NMutInfk\ measures the fraction of label entropy explained by the top-$k$ highlights (higher is better; e.g., $\NMutInfk{=}0.20$ corresponds to a $20\%$ uncertainty reduction). We invert the $y$-axis for \MDLBitsk\ so that higher values are better. In Fig.~\ref{fig:topk-mutual-information-and-mdl-prob-conflicting-or-irrelevant}, \textit{\MDLBitsk\ and \NMutInfk\ show similar trends with the \DrankkGroup\ results in Fig.~\ref{fig:rank-topk-margin-conflicting-or-irrelevant}.}. In the Conflicting setup, \FA\ is typically the best but variable (using the top-$k$ explanation importance scores can reduce about 19.8\% uncertainty in model answer label prediction, for half of the model-dataset cases), and \MI\ is second best (about 16.5\% uncertainty reduction). Following are \IG\ and \Attn, leaving about 91\% of label uncertainty. In the Irrelevant setup, all methods improve on both metrics, with \MI\ showing better performance than \FA{}. This again indicates that explanations can more effectively reveal context usage when the context is off‑topic. As expected, \NMutInfk\ and \MDLBitsk\ show similar trends.

Overall, \textbf{\MI\ shows best performance regarding whether the model relied on CK or PK}, followed by \FA{}, but with considerable variability in performance.
\IG\ and \Attn\ provide little value for this purpose. We hypothesise that \FA\ exhibits high variance due to its discrete perturbation nature: token replacement can introduce distribution shift, and small perturbations can trigger nonlinear interactions that change the answer in unstable ways. Moreover, our implementation attributes only the \emph{first generated token}, which can be a noisy objective for multi-token answers; when many tokens have near-tied effects, the top-$k$ ranking becomes sensitive to small score fluctuations. These factors jointly make \FA\ both computationally expensive and less stable for token-level context attribution in long-context regimes.

\begin{figure*}[ht]
  \centering
  \begin{subfigure}[t]{0.48\textwidth}
    \centering
    \includegraphics[width=\textwidth]{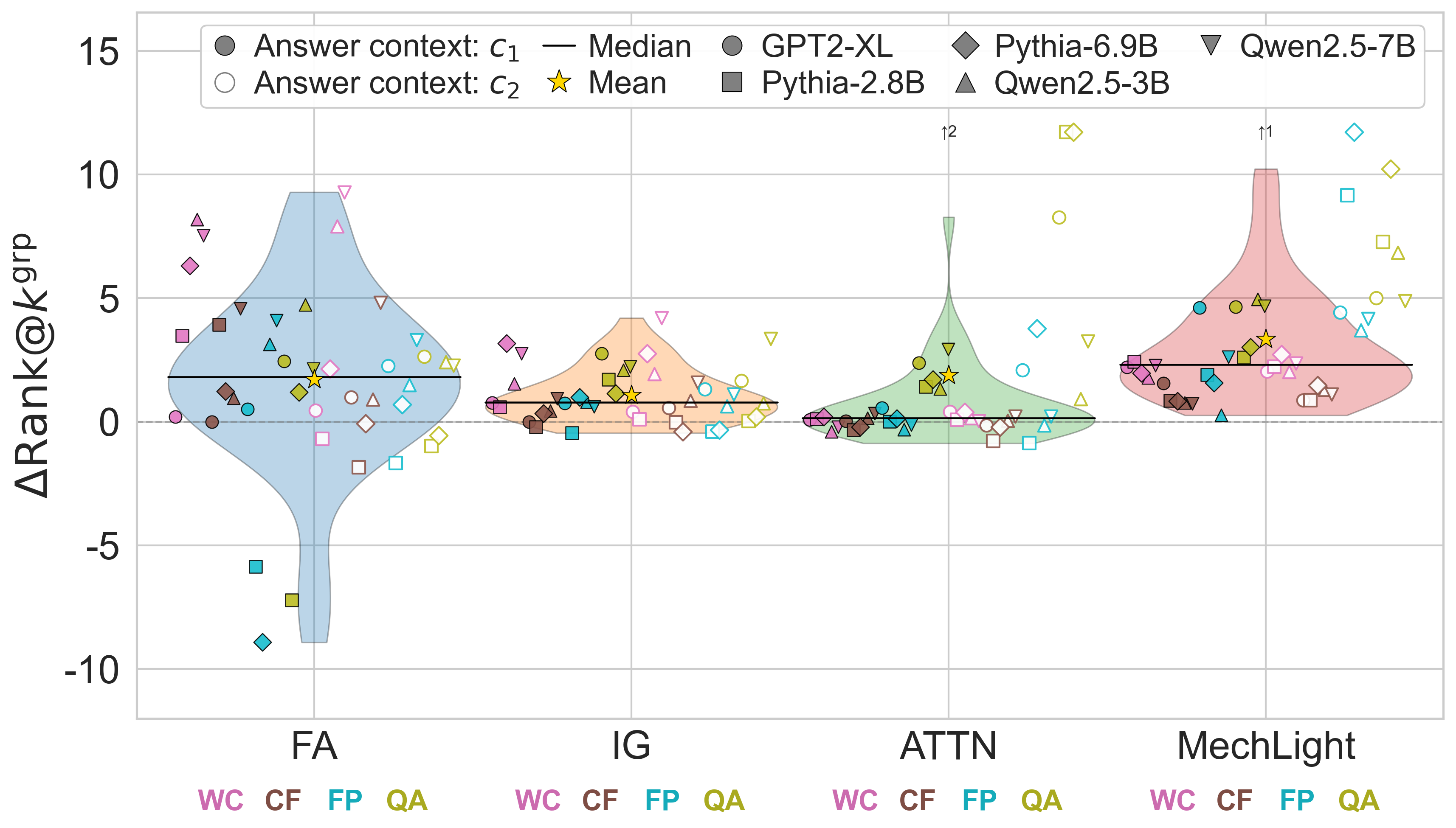}
    \caption{Double-Conflicting: Two Conflicting Contexts}
    \label{fig:conflicting-1-and-2-topk-5-rank-datasets-margin-only}
  \end{subfigure}
  \hfill
  \begin{subfigure}[t]{0.48\textwidth}
    \centering
    \includegraphics[width=\textwidth]
    {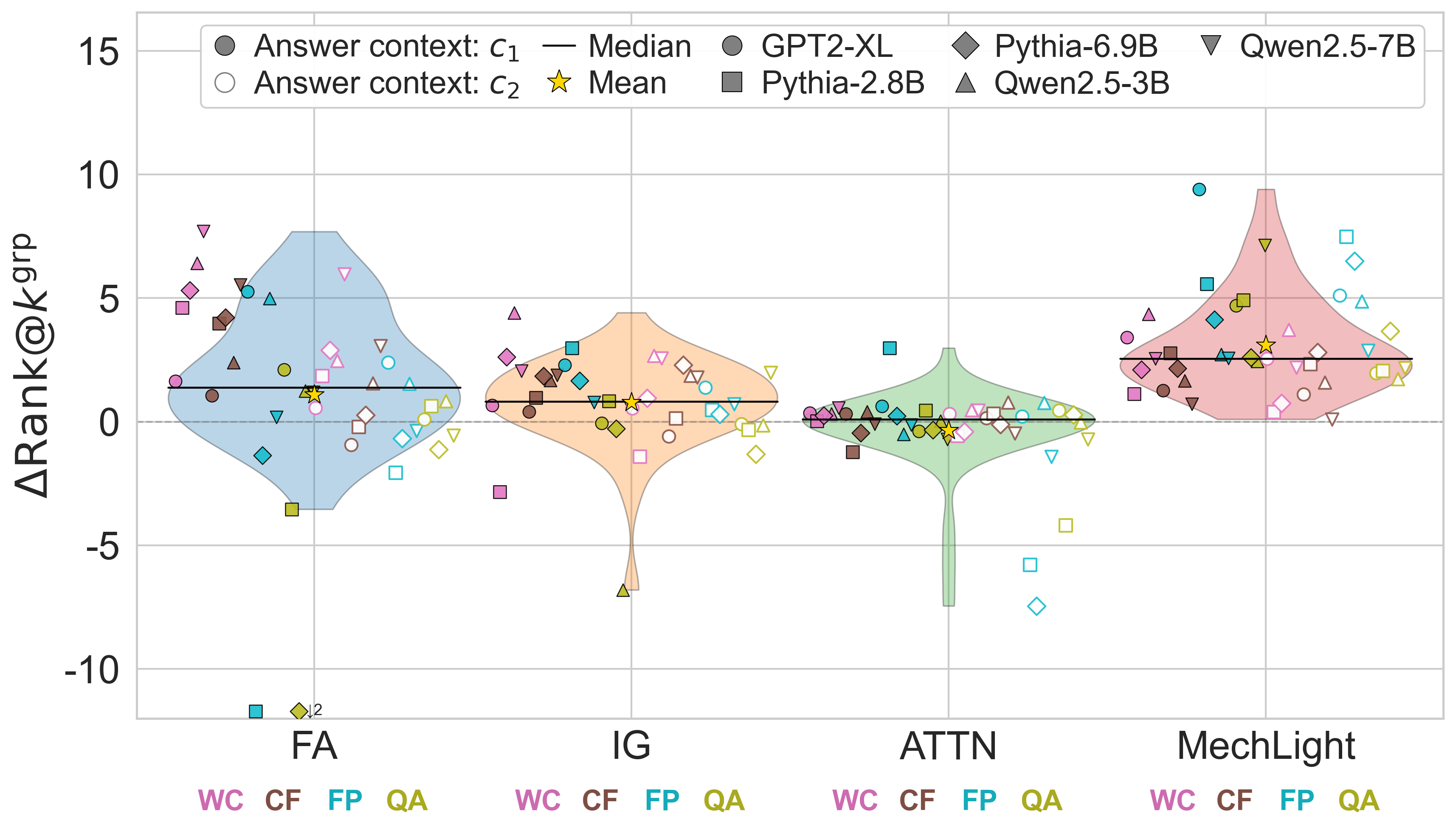}
    \caption{Mixed: One Irrelevant and One Conflicting Context}
    \label{fig:irrelevant-and-conflicting-topk-5-rank-datasets-margin-only}
  \end{subfigure}
  \caption{  
  \DrankkGroup\ (Eq.~\ref{eq:drankk-cross-instance-group}) -- average margins for the rank of context $c_1$ and $c_2$ between two instance groups $D_{c_1}$ and $D_{c_2}$  in the \textbf{Double-Conflicting} and \textbf{Mixed} setup (\S\ref{input-regime-2}). Higher \DrankkGroup is better, and a positive number means that the explanation can correctly indicate which context piece is used for the answer (\S\ref{sec:rq2-discussion}). The values are obtained by comparing the average importance scores of either the first context piece  ($c_1$, filled shapes at the left) or the second context piece ($c_2$, hollowed shapes at the right) between the two instance groups. The colors denote the datasets, and the marker shapes denote the models.
  }
  \label{fig:rank-topk-margin-double-conflicting-and-mixed}
\end{figure*}

\subsection{Does the explanation show \textit{which} of two context documents the model used?}
\label{sec:rq2-discussion}

\textbf{Document-level attribution across groups}. 
Fig.~\ref{fig:rank-topk-margin-double-conflicting-and-mixed} reports \DrankkGroup\ for dual-context setups. For each attribution method and model--dataset pair, we show two margins: filled markers (left half) correspond to instances where the model answers from $c_1$ (comparing $D_{C_1}$ vs.\ $D_{C_2}$), and hollow markers (right half) correspond to answers from $c_2$. Positive margins mean tokens in the utilised passage are ranked higher than those in the unused passage; thus, a good explanation method should have positive values in both halves.
We observe that \textit{\MI\ is the best in both setups, by consistently showing positive margins}, meaning that the answer-context tokens from the used context are actually ranked higher than tokens from the unused context. 
\textit{\FA\ is second-best overall, but often shows the largest negatives on long-context datasets (Fakepedia, ConflictQA) when the model answers from $c_1$ (filled markers)}, suggesting that \FA\ tends to prefer the later passage in concatenated prompts in these cases.
Following are \IG\ and \Attn\, with most margins close to zero in both setups, indicating they often fail to indicate which document the model selects the answer from. Comparing setups, the results are similar; \FA\ and \IG\ show slightly larger margins in Double-Conflicting and more variability in Mixed, with worse results on Fakepedia and ConflictQA, likely reflecting their sensitivity to context length and difficulty with long mixed contexts. The fact that \FA\ and \MI\ are better than \IG\ and \Attn\ again confirms the potential link between the faithfulness and the explanation utility (See faithfulness in Tab.~\ref{tab:faithfulness} in App.~\ref{app:additional_results}). Trends persist after swapping the two context pieces, in Double-Conflicting-Swap and Mixed-Swap (see Fig.~\ref{fig:rank-topk-margin-double-conflicting-and-mixed-after-swap} in App.~\ref{app:additional_results}).

\begin{figure*}[ht]
  \centering
   \begin{subfigure}[t]{0.48\textwidth}
    \centering
    \includegraphics[width=\textwidth]
    {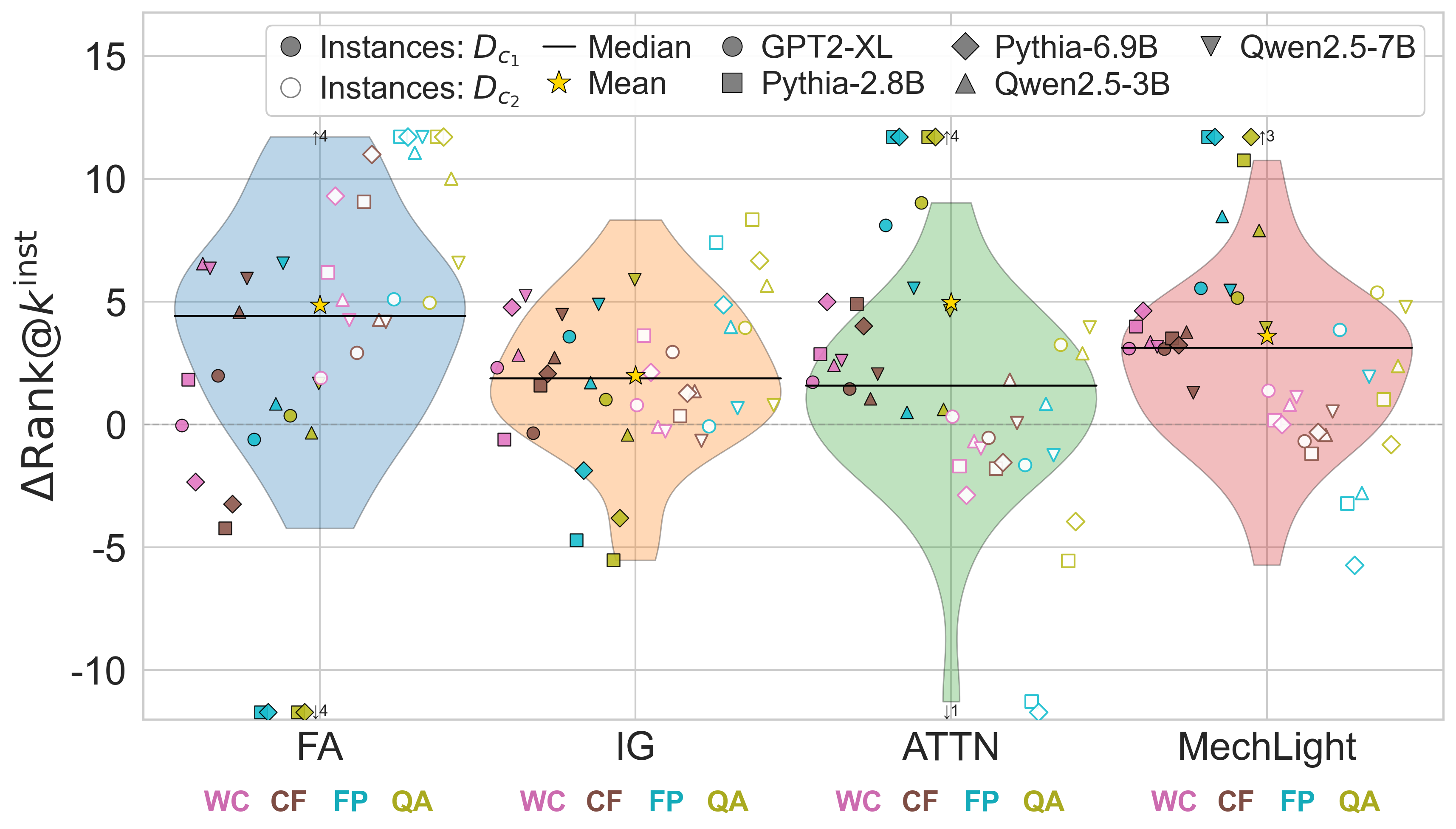}
    \caption{Double-Conflicting: Two Conflicting Contexts}
    \label{fig:conflicting-1-and-2-topk-5-rank-datasets-within-group-margin-only}
  \end{subfigure}
  \hfill
  \begin{subfigure}[t]{0.48\textwidth}
    \centering
    \includegraphics[width=\textwidth]
    {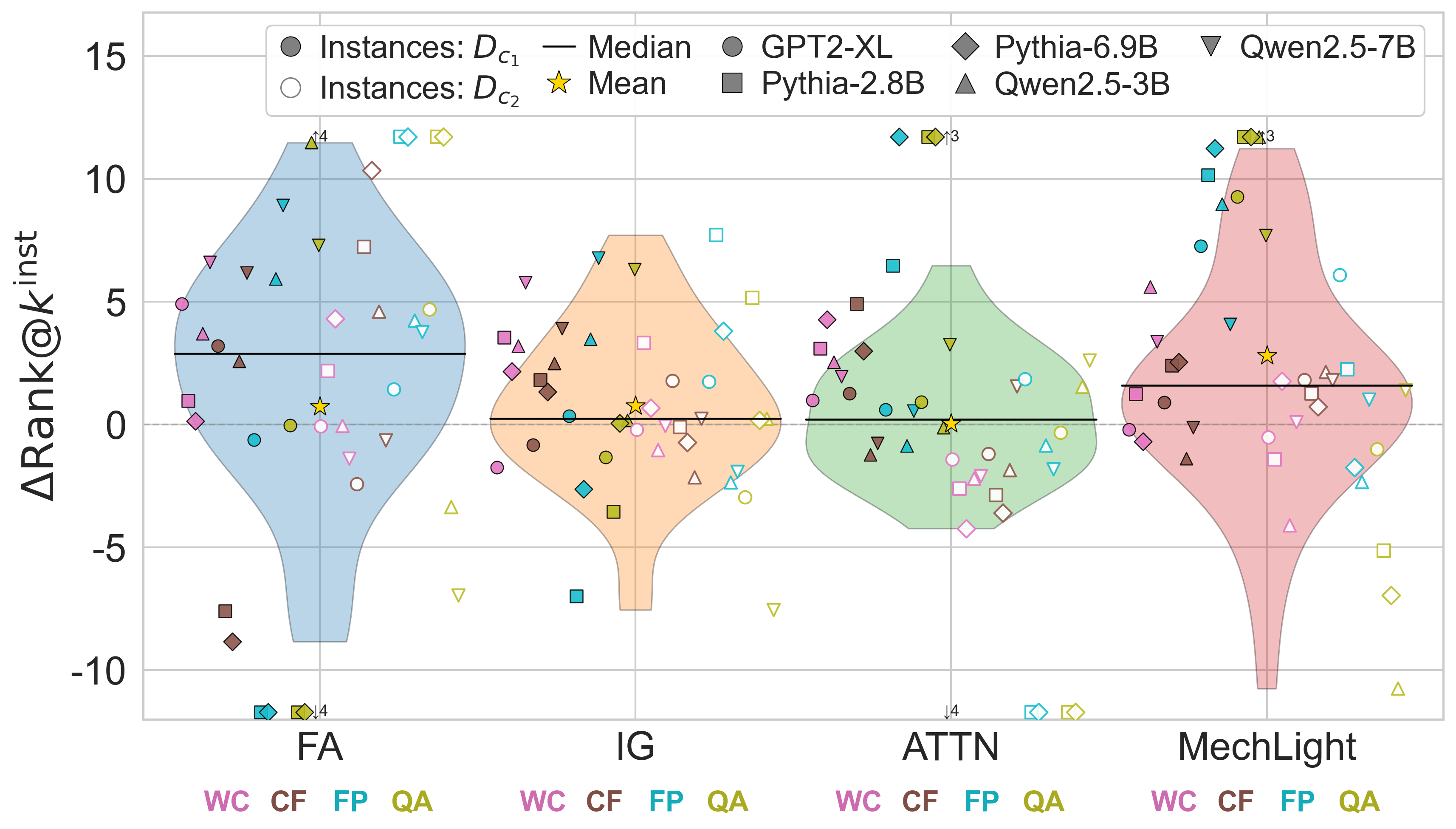}
    \caption{Mixed: One Irrelevant and One Conflicting Context}
    \label{fig:irrelevant-and-conflicting-topk-5-rank-datasets-within-group-margin-only}
  \end{subfigure}
  \caption{\DrankkInst\ (Eq.~\ref{eq:drankk-within-instance})  -- average \emph{within-instance-group} margins between the rank of the answer context piece and the other context piece in the \textbf{Double-Conflicting} and \textbf{Mixed} setup (\S\ref{input-regime-2}).  Higher \DrankkInst\ is better, and a positive number means that the explanation can correctly indicate which context piece is used for the answer (\S\ref{sec:rq2-discussion}). The values are obtained by comparing importance scores of the answer-piece context and the other piece within either the instance group answering from the first context piece  ($D_1$, filled shapes at the left) or that answering from the second context piece ($D_1$, hollowed shapes at the right). The colors denote the datasets, and the marker shapes denote the models.}
  \label{fig:rank-topk-margin-within-instance-double-conflicting-and-mixed}
\end{figure*}

\textbf{Document-level attribution across instances}. We now compare per instance the top‑$k$ rank margin between tokens in the utilised vs.\ unused document. Concretely, \DrankkInst\ (Fig.~\ref{fig:rank-topk-margin-within-instance-double-conflicting-and-mixed}) is positive when, within the same instance, tokens in the utilised passage outrank tokens in the unused passage according to the HE scores. As shown in Fig.~\ref{fig:rank-topk-margin-within-instance-double-conflicting-and-mixed}, no HE shows positive margins for all cases, especially on long contexts (Fakepedia and ConflictQA), implying \textit{the HEs often cannot indicate which document the answer is selected from, especially when the contexts are relatively long}. \MI\ is strongest overall (best in (b), second‑best in (a)) with positive rank margins in most cases. \FA\ follows, \IG\ and \Attn\ exhibit minor positive margins. We also find that all HEs exhibit positional bias: margins turn negative when the answer comes from the second (\MI, \Attn, which are based on the attention head mechanism) or first (\FA, \IG) piece in long contexts. The same trends hold in both setups and persist after changing piece order (Fig.~\ref{fig:rank-topk-margin-within-instance-double-conflicting-and-mixed-after-swap}), confirming the content-independent positional bias. We hypothesize that this opposing positional bias could either stem from the model structure or the nature of the attribution method (See discussion in \S\ref{sec:limitation} Limitation).

\begin{figure*}[ht]
  \centering
  \begin{subfigure}[t]{0.48\textwidth}
    \centering
    \includegraphics[width=\textwidth]
    {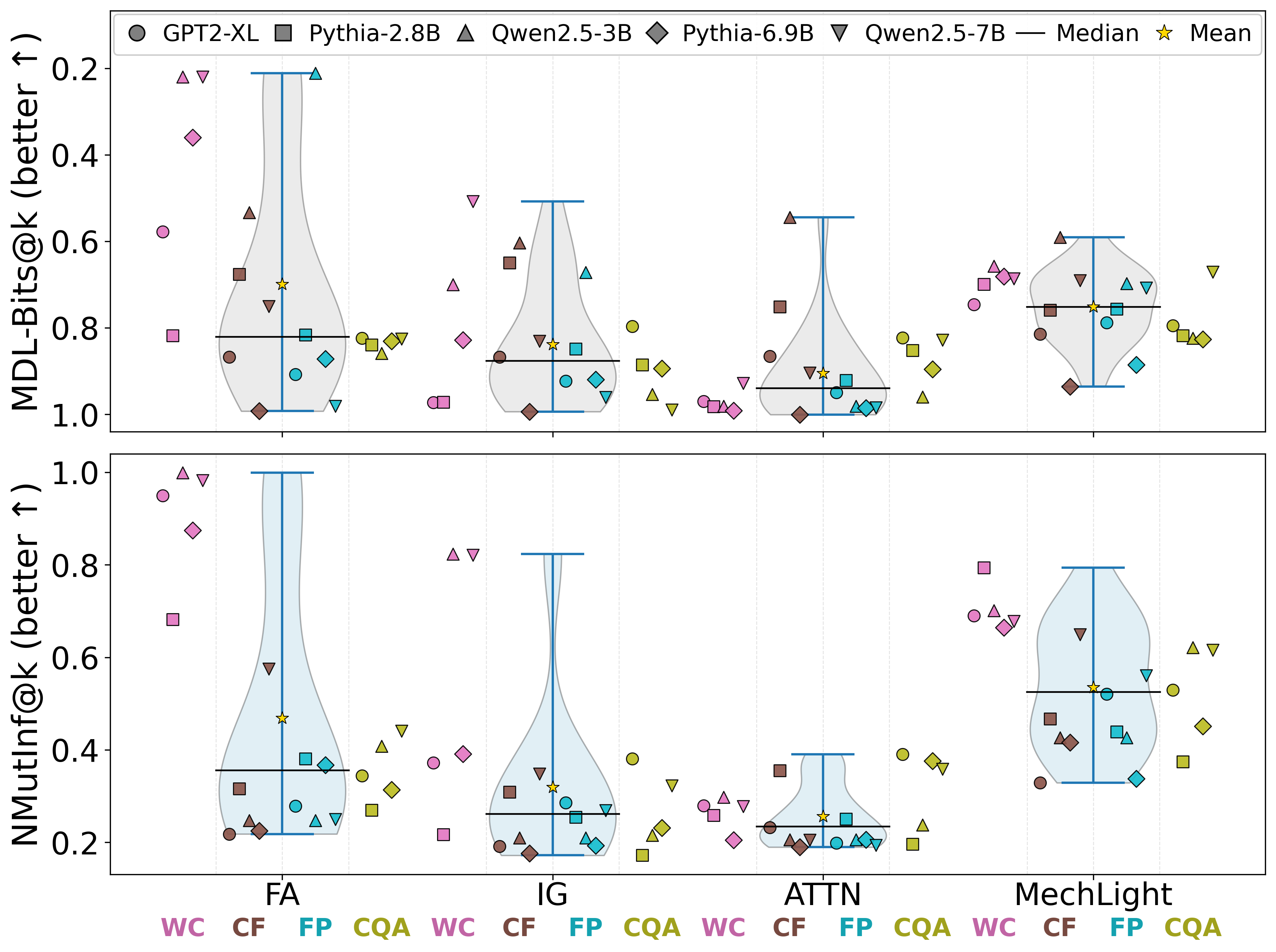}
    \caption{Double-Conflicting: Two Conflicting Contexts}
    \label{fig:conflicting-1-and-2-topk-5-mutual-information-and-mdl-prob}
  \end{subfigure}
  \hfill
  \begin{subfigure}[t]{0.48\textwidth}
    \centering
    \includegraphics[width=\textwidth]
    {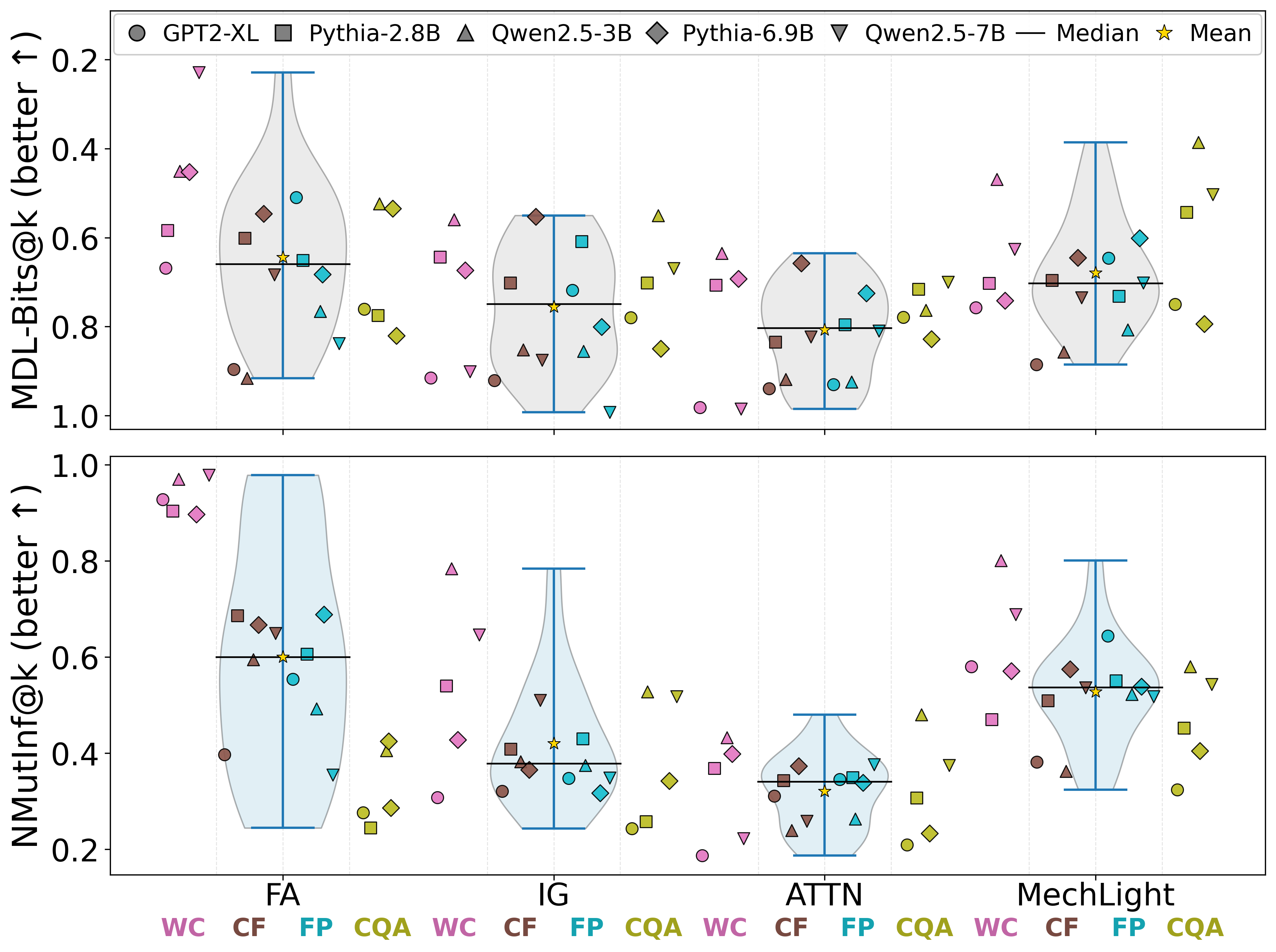}
    \caption{Mixed: One Irrelevant and One Conflicting Context}
    \label{fig:irrelevant-and-conflicting-topk-5-mutual-information-and-mdl-prob}
  \end{subfigure}
\caption{\MDLBitsk\ (top; y-axis is inverted; Eq.~\ref{eq:mdl}) and \NMutInfk\ (bottom; Eq.~\ref{eq:mutual-information}) in \textbf{Double-Conflicting} and \textbf{Mixed} setups (\S\ref{sec:four-context-setups}). Both \MDLBitsk (smaller value) and \NMutInfk are higher the better, indicating that the explanation importance scores are better at revealing the model's answer selection across the two context passages, i.e., from the first vs from the second piece. (\S\ref{sec:rq2-discussion}). The colors denote the datasets, and the marker shapes denote the models.
  }
    \label{fig:topk-mutual-information-and-mdl-prob-double-conflicting-or-mixed}
\end{figure*}

\textbf{Simulatability}. In  Fig.~\ref{fig:topk-mutual-information-and-mdl-prob-double-conflicting-or-mixed}, \MDLBitsk\ and \NMutInfk\ support the document-level attribution evaluation across groups. \textit{\MI\ is the best overall}, leading to $24.9\%$ uncertainty reduction on the label prediction given the top-$k$ highlights. \textit{Following is \FA{}}, which removes about $17.9\%$ of the label prediction uncertainty, \textit{but again with a variable performance}. 
\IG\ and \Attn\ show worse performance 
leaving most label prediction uncertainty.
Comparing the two input regimes, Double-Conflicting and Mixed, the findings are overall consistent and persist after position swapping of the two contexts (See Fig.~\ref{fig:topk-mutual-information-and-mdl-prob-double-conflicting-or-mixed-swapped} in App.~\ref{app:additional_results})



\begin{figure*}[ht]
  \centering
  \begin{subfigure}[t]{0.48\textwidth}
    \centering
    \includegraphics[width=\textwidth]
    {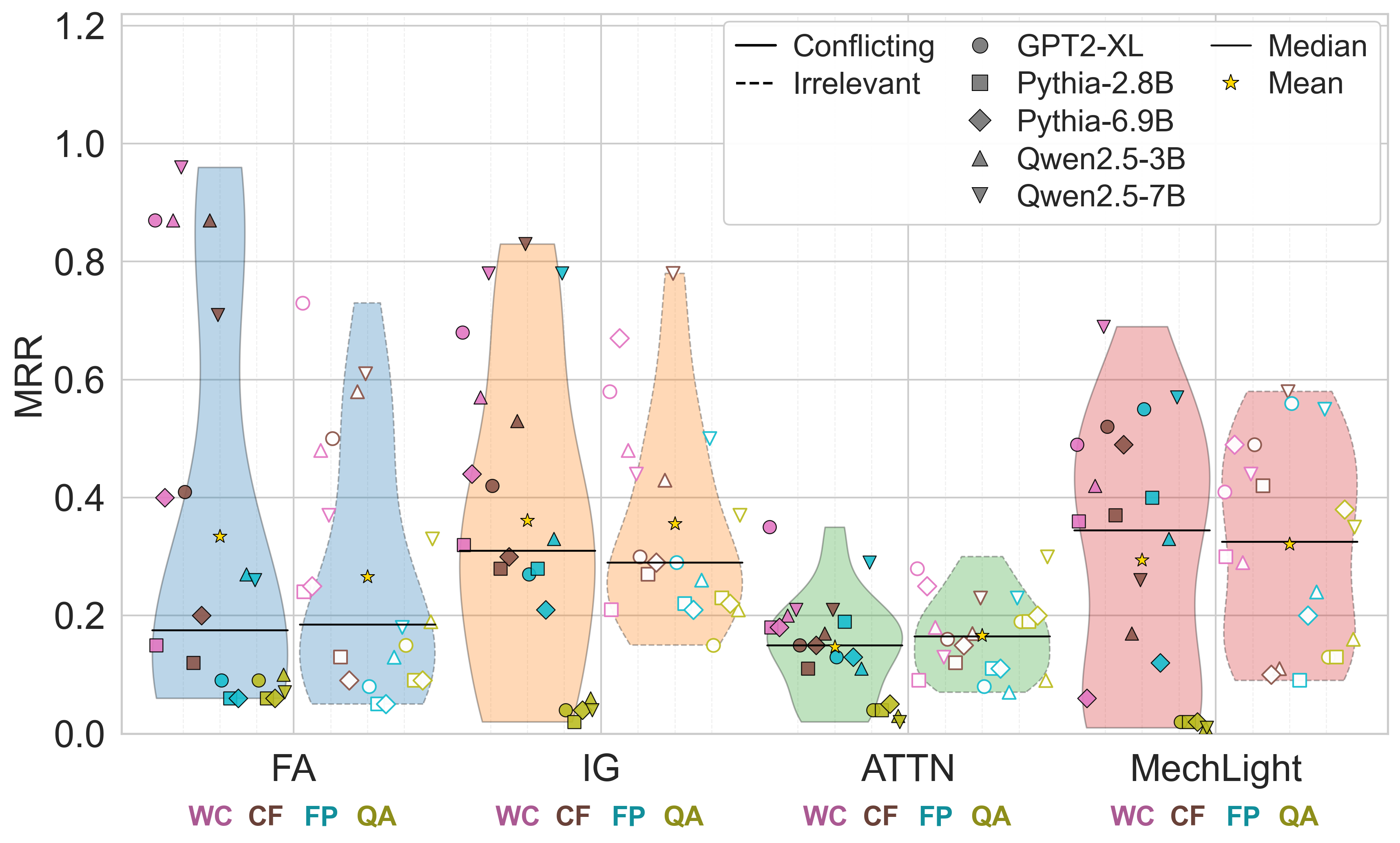}
    \caption{Conflicting \& Irrelevant}
    \label{fig:answer-token-mrr-conflicting-or-irrelevant}
  \end{subfigure}
  \hfill
  \begin{subfigure}[t]{0.48\textwidth}
    \centering
    \includegraphics[width=\textwidth]{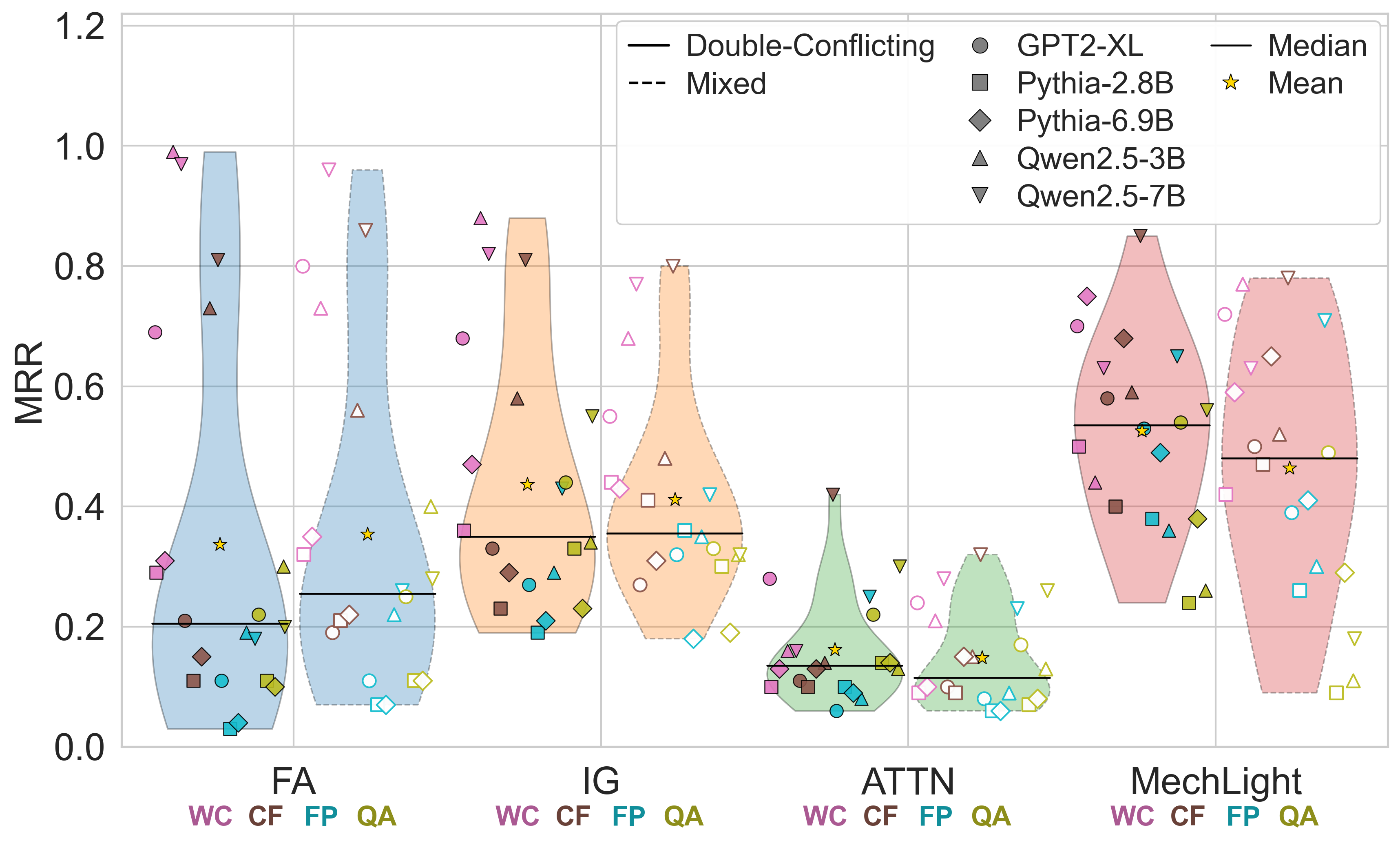}
    \caption{Double-Conflicting \& Mixed}
    \label{fig:answer-token-mrr-double-or-mixed}
  \end{subfigure}
  \caption{\mrr\ (Eq.~\ref{eq:mrr}) -- Mean Reciprocal Rank for the predicted answer tokens within the context-answer instances for all four context setups (\S\ref{input-regime-4}). Higher \mrr\ is better; larger values indicate the true answer token is placed near the top of the ranked list, meaning the explanation can more precisely indicate the answer location (\S\ref{sec:rq3-discussion}). Filled shapes stand for \textbf{Conflicting} setup and \textbf{Double-Conflicting} setup in the left and right subplot, respectively; hollow shapes stand for \textbf{Irrelevant} setup and \textbf{Mixed} setup in the left and right subplot, respectively. The colors denote the datasets, and the marker shapes denote the models.}

  \label{fig:answer-token-mrr-conflicting-irrelevant-double-mixed}
\end{figure*}

\subsection{Does the explanation pinpoint the exact context part(s) that were employed for the generated answer?}
\label{sec:rq3-discussion}
We report \mrr\ across models and datasets for all context setups in Fig.~\ref{fig:answer-token-mrr-conflicting-irrelevant-double-mixed}. Recall that \mrr\ is computed over context-answer instances, and larger values mean the gold answer token is ranked closer to the top (e.g., \mrr{}=$0.1$ corresponds to roughly rank $10$ on average).
From Fig.~\ref{fig:answer-token-mrr-conflicting-irrelevant-double-mixed}, we observe that \textit{all methods except \Attn\ usually place the answer token within the top‑10 ranks for most model–dataset combinations}. This is reflected by medians above $\approx 0.1$ for \MI/\IG/\FA\ in most model--dataset pairs, whereas \Attn\ often has lower values. \MI\ is the best performing, although its performance lowers on the long‑context dataset \textsc{ConflictQA}.

When a single piece of context is supplied (e.g., the \emph{Conflicting} context), as shown in Fig.~\ref{fig:answer-token-mrr-conflicting-or-irrelevant}, \textit{\MI\ and \IG\ are the two best methods} (median \mrr\ of 0.345 and 0.310, respectively), implying that the HEs often position the answer token within the top-3 tokens. \FA\ is next, with a median \mrr\ 0.175, but once again exhibits the largest variability between models and datasets
and low \mrr\ in long context datasets, suggesting that \FA\ is unstable and could require a computationally prohibitive number of ablations on long-context datasets. 
\Attn\ performs worst with a mean 0.147 \mrr.
\textit{As the context length increases (ConflictQA), all explanations struggle to position the answer tokens even within the top 10 important tokens.} Similar trend is found in Irrelevant setup, all methods show lower \mrr\ on short‑context datasets (World Capital, Counterfact) and slightly higher on long contexts (notably \textsc{ConflictQA}), suggesting that explanations are easily distracted by short, irrelevant information.

With two pieces of context, \MI\ performs best, with an average \mrr\ of 0.526, followed by \IG\ (average \mrr\ 0.436). \FA\ again shows the highest variability and performs poorly on long‑context datasets (e.g., Fakepedia), where answer tokens usually fall outside the top-10 most important tokens. \Attn\ remains consistently worst, with an average \mrr\ 0.162. All methods show similar but slightly lower \mrr\ in the Mixed Context setup. Trends hold after swapping the two contexts in Double-Conflicting-Swap and Mixed-Swap (Fig.~\ref{fig:answer-token-mrr-double-or-mixed-after-swap} in App.~\ref{app:additional_results}), indicating that the relative position of the context does not affect the overall utility of the explanations in locating the tokens of the answer in the context.

\textbf{Case study.}
To better understand why HEs often fail to pinpoint the exact answer span, we conduct a qualitative case study in the \textit{Conflicting} regime. We define a failure as an instance where the model copies the context answer, yet none of the top-$k{=}5$ attribution tokens match the answer token. Inspecting 50 randomly sampled failures per dataset (World Capital, CounterFact, ConflictQA, Fakepedia), we find three recurring patterns  (examples in App.~\ref{app:case_study}, Tab.~\ref{tab:failure-examples}). (\textit{i}) \textbf{Generic tokens dominate:} on shorter contexts (World Capital, CounterFact), failures often place high attribution on stop-words, punctuation, or boiler-plate openers (e.g., ``The'', ``Fact:''). (\textit{ii}) \textbf{Nearby descriptors/entities:} on longer contexts (ConflictQA, Fakepedia), explanations frequently highlight nearby descriptors or entity names adjacent to the gold answer while missing the answer token itself. (\textit{iii}) \textbf{Question focus:} across datasets, attributions can concentrate on the interrogative phrase (e.g., ``What'', ``Who'', ``In what country''), which provides little evidence about \emph{where} the answer was sourced in the context. Together, these patterns help explain why token-level localisation can fail even when the model’s answer is copied from context.

\section{Conclusion}
\label{sec:conclusion}
We introduce the first gold standard framework for evaluating highlight explanations (HEs) for context utilisation. It encompasses controlled test cases under known ground-truth context utilisation scenarios, enabling direct assessment of HE accuracy in context attribution.
Across four controlled context scenarios, five models, and four datasets, we demonstrate our framework's general applicability using three established HE methods and one mechanistic interpretability-based method (\MI{}). We find that \MI\ shows the highest utility across all context scenarios and that some commonly used HE methods, \IG\ and \Attn, provide no value in making context usage transparent. Furthermore, all methods suffer from long contexts and exhibit position bias when two contexts are provided. This calls for future highlight explanation methods to provide accurate and reliable explanations of context usage at scale. 

\section{Limitations}
\label{sec:limitation}
Our work introduces the first benchmark robustly evaluating HEs for context-usage utility. Here, we discuss its scope and opportunities for extension.

\textbf{Input regimes.}
Our four input context setups all ensure each answer can be traced to \emph{one} dominant source (CK, PK, or one of two passages). Interesting future extensions are tasks requiring \emph{joint} reasoning over multiple passages (e.g., multi‑hop QA or document‑level summaries), where saliency must reflect blended evidence. In such settings, for example in multi-hop QA where the answer depends on multiple supporting contexts and evidence spans, the framework could be extended from single-source to set-valued gold evidence, provided that the gold contexts and supporting spans can first be identified. RQ1 could still evaluate whether the model relied on contextual rather than parametric knowledge, while RQ2 would extend to multiple gold context pieces and RQ3 to multiple gold supporting spans. This would require set-valued extensions of our metrics (\S\ref{sec:metrics}); for example, document-level attribution could compare the importance assigned to the union of tokens from all gold contexts against unused contexts, while token-level attribution could evaluate the average rank of multiple gold spans rather than a single token. The research questions could also be made more fine-grained, e.g., by measuring how many required context pieces are identified by the HEs and what proportion of the supporting spans they localise. We leave these extensions to future work. Notably, we specifically focus on cases where the answer in the provided context contradicts the model's parametric knowledge, as this is where the model is ensured to utilise the context information, and thus the HEs are expected to highlight the context. Additionally, such a gold standard for the model's context utilisation in PK-aligned (supportive) contexts or blended CK/PK behaviour~\citep{cheng2024understanding} is missing; thus, it remains an open challenge to evaluate the utility of highlight explanations in such setups. 

\textbf{Dataset and output scope.} 
We target QA datasets with \emph{present and short} gold answer spans in the context, enabling the development of our gold standard assessment of HE accuracy for context utilisation tasks. These tasks enable gold supervision for both (i) source attribution (CK vs.\ PK, RQ1, or which passage, RQ2) and (ii) token-level golden answer localisation (RQ3). However, these assumptions may not hold in open-ended or long-form generation, where outputs can lack a unique answer string that maps cleanly to a single supporting span. A possible extension is to decompose the generated output into sentences and infer, for each output sentence, its most supported source context (e.g., via entailment-style scoring or an LLM judge), yielding pseudo-gold source labels for the model's context usage source for studying RQ1/RQ2. For RQ3, one could align each sentence back to one or more supporting spans in the selected source, producing a set of reference spans rather than a single span. Finally, since \FA, \IG, \Attn, and \MI\ are usually defined per decoding step, applying these explanations to long-form settings would require aggregating attributions across multiple generated tokens. This would require additional design choices to avoid aggregation noise for attribution scores. Thus, developing robust pseudo-gold construction and attribution aggregation methods are promising directions to apply the framework to long-form generation tasks in the future.

\textbf{Explaining reasoning trajectories.} Our framework only focuses on short-answer QA tasks, and the explainability techniques compute attributions at a single decoding position (the answer token). However, reasoning models may produce an explicit multi-step \emph{reasoning trace} (a sequence of intermediate steps) before the final answer. An extension of our framework could compute token attributions over the input at multiple decoding steps, and then aggregate these attributions within each reasoning step (e.g., by sentence) to obtain highlight explanations for each reasoning step. This would enable step-wise versions of RQ1--RQ3, allowing us to study how CK-vs-PK reliance, passage preference, and evidence localisation shown by highlight explanations evolve over the course of reasoning.


\textbf{Interpreting opposing positional biases in multi-document setups.}
In dual-context setups, we observe opposing positional preferences across explanation families (\S\ref{sec:rq2-discussion}), and swapped-passage variants suggest the effect is largely position-driven rather than purely content-driven. We conjecture two contributing factors.

\textbf{(i) Model-induced position effects.}
Long-context studies show that whether evidence affects the final prediction depends strongly on its position in the prompt \citep{liu-etal-2024-lost}, and this reliance can decay with distance from the prediction \citep{khandelwal-etal-2018-sharp}. In our experiments, the strongest biases occur for the Pythia models (Fig.~\ref{fig:rank-topk-margin-double-conflicting-and-mixed}), suggesting that model architecture may modulate position sensitivity. One plausible contributor is positional encoding: RoPE encodes position via rotations and yields distance-dependent interactions \citep{su-etal-2021-roformer}, and recent theory suggests that RoPE encodings interacting with causal masking can produce primacy--recency trade-offs \citep{wu-etal-2025-position-bias}.

\textbf{(ii) Method-induced measurement bias.}
\IG/\FA\ are sensitivity-based: they quantify how the next-token distribution changes with gradients or input perturbations; their outcomes can depend on baseline choices, and perturbations/removals can also introduce distribution shift \citep{lyu-etal-2024-towards}. In contrast, \Attn\ and \MI\ are attention-based: they reflect attention distributions; however, attention weights are not guaranteed to track causal contribution \citep{jain-wallace-2019-attention,wiegreffe-pinter-2019-attention,serrano-smith-2019-attention}. Moreover, local residual-stream readouts ( used by \MI{}) can be confounded when later components erase or cancel earlier residual directions \citep{janiak-etal-2024-adversarial,belrose2023tunedlens}. These measurement differences may contribute to the opposing positional biases observed across methods.

Overall, the observed positional preference of an explanation should be interpreted with caution: it may reflect genuine model position sensitivity, explanation method-induced artifacts, or both. We leave disentangling these factors to future work.

\textbf{Model scale and architecture.}
Our experiments systematically cover five models up to 7B parameters and reveal HE accuracy shifts with context length and model scale. While these results establish clear trends in this setting, extending the framework to instruction-tuned or much larger models may reveal additional context-usage patterns across all three research questions. For RQ1, instruction tuning may affect how strongly models prioritise provided context over parametric knowledge, as context use can become more sensitive to task directives and prompt framing. For RQ2, instruction tuning or larger scale may also affect how models arbitrate between multiple passages, potentially changing the passage-selection and positional-bias patterns observed in our dual-context setups. For RQ3, much larger models may rely on more distributed mechanisms for extracting answer evidence, which could make single-head projections less complete and motivate HE extraction methods that go beyond a single selected head. Future work could therefore compare matched variants with and without instruction tuning, as well as larger scales, to test whether the CK/PK separation, multi-passage selection, and answer-span localisation patterns observed here persist or change under these regimes.

\textbf{Explanation families.}
Our benchmark spans three standard post-hoc techniques plus our novel MI-based method. The framework's flexible architecture enables seamless integration of additional HE variants, both post-hoc and MI, for future investigation. \MI\ is designed to be lightweight and to produce token-level highlights from a single selected attention head, but this simplicity comes with limitations. First, the direct-logit readout provides a local attribution signal: later layers may reinforce or counteract earlier residual directions (downstream cancellation) \citep{janiak-etal-2024-adversarial,belrose2023tunedlens}, so a head’s direct contribution to the candidate logit gap may not fully reflect its end-to-end causal effect on the final answer. Second, \MI\ uses attention weights as the highlight object; attention is not guaranteed to reflect causal information flow~\citep{jain2019attention,wiegreffe2019attention}. A promising direction for future work is to integrate more intervention-based signals, e.g., causal head ranking via ablation~\citep{yu-etal-2023-characterizing,jin-etal-2024-cutting}, to improve causal faithfulness, and to explore patching- or neuron-centric mechanisms~\citep{meng2022locating,wang2023detecting,wang2024wh,shi2024ircan} to capture distributed context-usage circuits, while designing principled projections of such component-level evidence into stable token-level highlights.


\textbf{Explanation utility and human perspective.} Our framework leverages automated gold standard metrics, uniquely enabled by context usage scenarios where ground-truth source attribution is known. Supplementary faithfulness analyses validate these findings. While our principled automated approach avoids annotation costs, future human studies remain valuable for assessing perceived utility. These design choices establish a rigorous foundation for context-usage HE evaluation, with clear pathways for extending to more complex scenarios and explanation paradigms.

\section*{Acknowledgements}
$\begin{array}{l}\includegraphics[width=1cm]{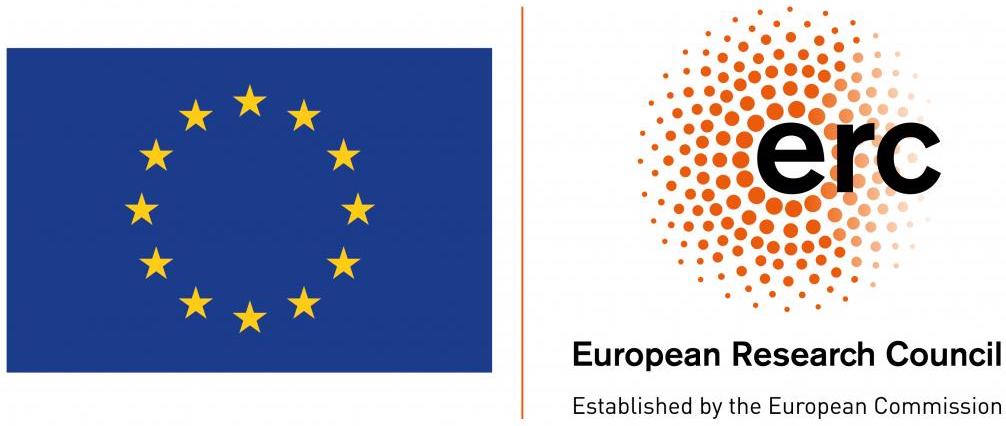} \end{array}$ 
This research was co-funded by the European Union (ERC, ExplainYourself, 101077481) and by the VILLUM FONDEN (grant number 40543). Views and opinions expressed are however, those of the author(s) only and do not necessarily reflect those of the European Union or the European Research Council. Neither the European Union nor the granting authority can be held responsible for them. 

\bibliographystyle{compling}

\bibliography{custom}
\appendix

\renewcommand{\theHsection}{app.\Alph{section}}
\renewcommand{\theHsubsection}{app.\Alph{section}.\arabic{subsection}}

\section{Replication Details}

\subsection{Datasets Details}
\label{app:dataset-details}
\textbf{Reconstruction overview.}
For each question, we construct matched instances across all regimes with token‑level supervision while keeping the question fixed:
\begin{enumerate}[nosep,leftmargin=1.25em]
  \item \textbf{Memory check.} Query the target model without context to obtain its parametric answer; retain only items whose “conflicting” contexts genuinely contradict that answer (drop candidates that leak the model's parametric answer).
  \item \textbf{Regime assembly.} Build \textsc{Conflicting}, \textsc{Irrelevant}, \textsc{Double‑Conflicting}, and \textsc{Mixed} prompts by concatenating passages so that each piece contains an explicit \emph{candidate answer token} (enabling RQ3).
  \item \textbf{Swaps} Create swapped dual‑context variants to control for position.
\end{enumerate}
This yields per‑question, per‑regime test sets with known gold spans and answer locations suited to our utility‑focused metrics. Dataset-specific construction details are as follows.

\textbf{Dataset-specific notes.}
\begin{itemize}[leftmargin=1.3em,itemsep=2pt]
  \item \textsc{Fakepedia}\; It contains encyclopaedic, single-hop questions spanning 45 Wikidata-style relations (e.g., \textit{employed-by}, \textit{official-language}).  The synthetic counterfactual context shipped with each item serves as the \emph{conflicting} context; an \emph{irrelevant} context is sampled from a different country that shares the same relation. See Table~\ref{tab:exmaple-for-fakepedia-dataset}.

  \item \textsc{World Capital}\;It contains purely geographical questions under a single relation, \textit{capital-of}.   The made-up capital statement is reused as the \emph{conflicting} context; an \emph{irrelevant} context is taken from another country. 

  \item \textsc{Counterfact}\;It contains entity-centric biography questions covering 5 relations such as \textit{works-in-area-of} and \textit{originated-in}. The dataset's edited context is kept as \emph{conflicting}; its annotated irrelevant context is reused. 

  \item \textsc{ConflictQA}\;It contains multi-domain questions across 7 relations (e.g., \textit{occupation},\textit{genre}, \textit{founded-year}).  The original contradictory context remains \emph{conflicting}; the supplied noise context (same relation, different subject) becomes \emph{irrelevant} after we extract the answer entity within the irrelevant context via Llama-4.  
\end{itemize}

\begin{table}[t]
\centering
\footnotesize
\setlength{\tabcolsep}{4pt}       
\renewcommand{\arraystretch}{0.92} 
\begin{adjustbox}{max width=\linewidth} 
\begin{tabular}{l l r r}
\toprule
Dataset & Ctx. type & \#Inst. & Avg ctx len \\
\midrule
\multirow{4}{*}{World Capital} & Conf.       & 55,830 &   37.9 \\
                               & Irre.       & 55,830 &   37.9 \\
                               & DoubleConf. & 55,830 &   75.9 \\
                               & Mixed       & 55,830 &   75.9 \\
\midrule
\multirow{4}{*}{Counterfact}   & Conf.       &    802 &   44.8 \\
                               & Irre.       &    802 &   44.8 \\
                               & DoubleConf. &    802 &   89.5 \\
                               & Mixed       &    802 &   89.5 \\
\midrule
\multirow{4}{*}{Fakepedia}     & Conf.       &  5,348 &  704.5 \\
                               & Irre.       &  5,348 &  704.5 \\
                               & DoubleConf. &  5,348 & 1408.8 \\
                               & Mixed       &  5,348 & 1408.9 \\
\midrule
\multirow{4}{*}{ConflictQA}    & Conf.       &  1,343 &  593.1 \\
                               & Irre.       &  1,343 &  454.1 \\
                               & DoubleConf. &  1,343 & 1190.2 \\
                               & Mixed       &  1,343 & 1047.2 \\
\bottomrule
\end{tabular}
\end{adjustbox}
\caption{Counts and average context length for reconstructed datasets regarding all four input regimes: Conf.(conflicting context); Irre.(irrelevant context); DoubleConf. (Double-Conflicting) contexts; Mixed (concatenation of conflicting and irrelevant contexts). Double-Conflicting-Swap and Mixed-Swap have identical statistics as DoubleConf. and Mixed (only positions are reversed).}
\label{tab:dataset-statistics}
\end{table}

Tab.~\ref{tab:dataset-statistics} summarises the statistics of the reconstructed datasets. To keep computation tractable, we cap the number of instances used for explanation generation and evaluation at 2,000 per dataset–context type for the short‑context datasets (World Capital, Counterfact) and 1,000 for the long‑context datasets (Fakepedia, ConflictQA), given the runtime overhead of Feature Ablation, which is more pronounced for long contexts.

\subsection{Other Details for Explanation Evaluation}
\label{app:k_3_and_9_for_rq1_and_2}
We select the top-$k$ important highlight explanations for utility evaluation, $k{=}5$ in the main discussion, as users often focus on a few instead of the complete cause of an event \citep{miller2019explanation}. To assess robustness, we conduct experiments with top-$3$ and top-$9$ explanations on a representative subset of regimes, as a human can usually hold $7\pm 2$ objects(here, explanation tokens) in short-term memory according to Miller's law\citep{baddeley1994magical}. The findings are consistent across different $k$.


\subsection{kNN Mutual Information Implementation Details}
\label{app:knnmi}
Given a top-$k$ highlight vector $X^{(k)}_s \in {R}^{k}$ extracted from a target segment $s$ (e.g., $s{=}c$ for RQ1 or $s\!\in\!\{c_1,c_2\}$ for RQ2) and a binary behaviour label $Y$ (RQ1: $\textsc{C}$ vs.\ $\textsc{M}$; RQ2: $\textsc{C1}$ vs.\ $\textsc{C2}$), we estimate the mutual information—i.e., the reduction in label uncertainty provided by the top-$k$ features—as
\begin{equation}
I\bigl(Y;X^{(k)}_s\bigr) \;=\; H(Y)\;-\;H\!\bigl(Y\mid X^{(k)}_s\bigr).
\end{equation}

\textbf{Label entropy.}
Let $p=\Pr(Y=1)$ be the empirical class prior. Using natural logarithms (nats),

\begin{equation}
H(Y)=
\begin{cases}
0, & p\in\{0,1\},\\
-p\log p-(1-p)\log(1-p), & p\in(0,1).
\end{cases}
\label{eq:hy}
\end{equation}

\textbf{kNN posterior and conditional entropy estimation.}
For each sample $x^{(k)}_{s,i}$, let $\mathcal{N}_k(i)$ be the set of its $k$ nearest neighbours in the feature space (Euclidean; the point itself is excluded; $k{=}5$). The local posterior (class–1 probability) is defined by the neighbour fraction:

\begin{equation}
\label{eq:post}
\begin{aligned}
\hat p_i
  &= \frac{1}{k}\sum_{j\in\mathcal{N}_k(i)} \mathbf{1}\{y_j=1\}\\
  &\approx \Pr\!\left(Y=1 \,\middle|\, X^{(k)}_s=x^{(k)}_{s,i}\right).
\end{aligned}
\end{equation}


With $h_b(q) = -q\log q -(1-q)\log(1-q)$ the binary entropy, the conditional entropy is estimated by averaging local entropies:
\begin{equation}
\widehat{H}\!\bigl(Y\mid X^{(k)}_s\bigr) \;=\; \frac{1}{n}\sum_{i=1}^n h_b(\hat{p}_i).
\label{eq:condent}
\end{equation}

\textbf{Normalised Mutual Information.}
The MI estimate is
\begin{equation}
\widehat{I}\bigl(Y;X^{(k)}_s\bigr) \;=\; H(Y)\;-\;\widehat{H}\!\bigl(Y\mid X^{(k)}_s\bigr).
\label{eq:mi}
\end{equation}
To express MI in bits, we use $\widehat{I}_{\text{bits}}=\widehat{I}/\log 2$.
Our reported quantity is the label‑entropy–normalised mutual information, i.e., the fraction of label uncertainty explained by the top‑$k$ highlights:
\begin{equation}
 \NMutInfk;=\; \frac{\widehat{I}\bigl(Y;X^{(k)}_s\bigr)}{H(Y)} \;\in\; [0,1].
\label{eq:nmi}
\end{equation}

\subsection{MDL Probe Implementation Details}
\label{app:mdl_prob}
In its classical formulation, the Minimum Description Length (MDL) principle provides a Bayesian-inspired framework for model selection. A model class $\mathbf{M}$ is a set of candidate models $M_i$; for example, $\mathbf{M}$ could be the family of cubic polynomials, with one member $M_i$ given by $5x^3$. Between two model classes $\mathbf{M}_a$ and $\mathbf{M}_b$, the preferred class is the one that yields the smaller \textit{stochastic complexity}, where the \emph{stochastic complexity} of $D$ with respect to a model class $\mathbf{M}$ is defined as the shortest achievable code length for $D$ when encoding is restricted to models in $\mathbf{M}$. Intuitively, a model that fits the data better assigns higher likelihoods and therefore produces shorter code lengths.

There are two standard methods for computing code lengths of deep neural nets. In the variational formulation \cite{hinton-mdl}, the description length of a dataset under a model is upper bounded by the sum of two terms: the negative log-likelihood of the data under the model and a complexity penalty given by the KL divergence between a variational posterior over parameters and a prior. 
This provides a tractable bound on stochastic complexity but depends strongly on the choice of prior and approximating family. Prequential (or online) coding measures description length by sequentially predicting the data. At each step, the model parameters are updated on past observations and used to predict the next outcome; the surprisal $-\log p(y_t \mid x_t, \theta_{t-1})$ is then added to the cumulative code length. The resulting quantity captures how efficiently a model class can compress data when trained incrementally. 
\citet{blier-mdl} shows that variational MDL often yields loose compression bounds, whereas prequential MDL produces much tighter estimates that align more closely with generalisation performance.

In NLP, MDL has been used in the context of ``probing tasks''. \citet{ep-tenney} used a suite of classifiers or probes to predict a token's syntactic (e.g., part-of-speech) and semantic tags from its embedding. A high accuracy in this task was interpreted as the embedding's ability to encode such linguistic information. The subsequent criticisms focused on the problem of ``classifier knowledge'' -- was the knowledge encoded in the embeddings, or did the classifier learn the task? \citet{voita-titov-2020-information} used ``MDL probing'' to solve this problem. Specifically, the prequential code lengths were computed using the formula $L_{\text{preq}}(\mathcal{D}) \;=\; \sum_{t=1}^{N} \ell_t \;=\; -\sum_{t=1}^{N} \log_2 p_{\theta_{t-1}}(y_t \mid h_t)$. Here $h=f_\phi(x)$ is a representation of a token from a frozen encoder $f_\phi$ and $p_\theta(y\mid h)$ is the predicted probabilities from a parametric probe. A lower $L_{\text{preq}}$ implied that the labels were easier to compress given the reps $h_t$, i.e., the property was more naturally encoded.

The MDL part of our simulatability test uses the same technique with top-$k$ importance scores derived from highlight explanations. We intend to show that these features have the discriminative power to predict a model's answer behaviour. We use a two-layer MLP classifier that is first trained on $10\%$ of the data. In the coding phase, we update the parameters $\theta$ for a mini-batch of size $10$. We repeat this entire process on $10$ random reshuffles of the data and report the average results.


\subsection{Faithfulness Evaluation Implementation Details}
\label{app:faithfulness}

Utility metrics in \S\ref{sec:methodology} assess how accurately a highlight explanation (HE) reflects the model's context usage. Faithfulness answers a complementary question: how well an HE aligns with the model's internal decision process. We therefore report \emph{Comprehensiveness} and \emph{Sufficiency} on the same models and datasets as the main experiments, under two regimes: \textbf{Conflicting} (single‑context) and \textbf{Double‑Conflicting} (dual‑context).

Following prior work~\citep{deyoung2019eraser,atanasova2021diagnosticsguided}, let $\pi(1{:}k)$ be the indices of the top‑$k$ tokens by HE scores $\phi$. For each $k\in\mathcal{K}$, let $x^{\mathrm{mask}\,k}$ be $x$ with tokens $\pi(1{:}k)$ masked, and $x^{\mathrm{keep}\,k}$ keep only $\pi(1{:}k)$. Writing $\ell(z)\!=\!\log p(a\mid z)$,
\begin{equation}\label{eq:aopc-comp}
\mathrm{AOPC}_{\text{comp}} \;=\; \frac{1}{|\mathcal{K}|}\sum_{k\in\mathcal{K}}\big[\ell(x)-\ell(x^{\mathrm{mask}\,k})\big],
\end{equation}
\begin{equation}\label{eq:aopc-suff}
\mathrm{AOPC}_{\text{suff}} \;=\; \frac{1}{|\mathcal{K}|}\sum_{k\in\mathcal{K}}\big[\ell(x)-\ell(x^{\mathrm{keep}\,k})\big].
\end{equation}
Higher $\mathrm{AOPC}_{\text{comp}}$ and lower $\mathrm{AOPC}_{\text{suff}}$ indicate greater faithfulness.

For World Capital/CounterFact (short contexts) we use $\mathcal{K}\!=\!\{1,\dots,5\}$. For Fakepedia/ConflictQA (long contexts), we use a fractional grid $\mathcal{K}\!=\!\{0.01,0.02,0.03,0.04,0.05\}\cdot n$ to avoid overly sparse inputs and keep the number of forward passes manageable.

\section{Additional Results}
\label{app:additional_results}

\subsection{Case Study for Explanation Failure}
\label{app:case_study}

To probe explanation failure for RQ3, whether a highlight explanation (HE) identifies the exact span supplying the model's answer, we focus on the \emph{conflicting‑context} regime. In this case study, we recognise a failure as when the model \emph{copies the correct answer from context} but \emph{none of the top-$k=5$ attribution tokens} matches the answer token. We examine 50 randomly sampled failures from each dataset (World Capital, ConflictQA, CounterFact, Fakepedia) and observe three recurring patterns. Tab.~\ref{tab:failure-examples} shows representative examples (top-$5$ highlight tokens underlined; prompt slices truncated with ellipses).

\begin{table*}[p]
\centering\footnotesize
\setlength{\tabcolsep}{5pt}
\renewcommand{\arraystretch}{1.08}
\begin{tabular*}{\textwidth}{@{\extracolsep{\fill}}
  >{\raggedright\arraybackslash}p{0.22\textwidth}
  >{\raggedright\arraybackslash}p{0.5\textwidth}
  >{\raggedright\arraybackslash}p{0.14\textwidth}
@{}}
\toprule
\textbf{Dataset} & \textbf{Prompt slice} & \textbf{Model answer} \\
\midrule
\multicolumn{3}{@{}l}{\textbf{Obs.\,1: Generic tokens (stop‑words / punctuation)}}\\
World Capital &
\underline{The} \underline{capital} \underline{of} Afghanistan \underline{is} \underline{.}\ Valletta \dots{} Q:\ What is the capital of Afghanistan? A: &
Valletta \\
World Capital &
\underline{The} \underline{capital} \underline{of} Algeria \underline{is} \underline{.}\ Sukhumi \dots{} Q:\ What is the capital of Algeria? A: &
Sukhumi \\
CounterFact &
\underline{Fact} \underline{:} Accra, \underline{the} \underline{capital} city \underline{of} Lebanon. Q:\ Accra, the capital city of A: &
Lebanon \\
\midrule
\multicolumn{3}{@{}l}{\textbf{Obs.\,2: Nearby descriptors/entity names}}\\
ConflictQA &
Trade paper profile begins \underline{Veteran} \underline{producer} \underline{and} \underline{studio} \underline{head} Dore Schary \dots{} Q:\ Who directed Act One? &
Dore Schary \\
ConflictQA &
Effects journal states \underline{Visual} \underline{effects} \underline{expert} Bruno \underline{was} \underline{hired} for ``Virus'' \dots{} Q:\ Who was the director of Virus? &
John Bruno \\
Fakepedia &
Apple Pay white‑paper: the \underline{ground} \underline{breaking} \underline{payment} \underline{service} \underline{launched} with Intel hardware \dots{} Q:\ Apple Pay, a product created by &
Intel \\
\midrule
\multicolumn{3}{@{}l}{\textbf{Obs.\,3: Question focus}}\\
World Capital &
\dots{} Q:\ \underline{What} \underline{is} \underline{the} \underline{capital} \underline{of} Albania? A: &
Berlin \\
ConflictQA &
\dots{} Q:\ \underline{Who} \underline{was} \underline{the} \underline{director} \underline{of} ``Virus''? A: &
John Bruno \\
CounterFact &
\dots{} Q:\ \underline{What} \underline{is} \underline{the} \underline{capital} \underline{of} Burgundy? A: &
Bangkok \\
\bottomrule
\end{tabular*}
\caption{Representative failure examples by pattern (top‑5 highlight tokens underlined).}
\label{tab:failure-examples}
\end{table*}


\textbf{Observations} 
(\textit{i}) \textbf{Spurious attention to generic words.} On short contexts (World Capital, CounterFact), about one‑third of failures place top‑5 highlights on punctuation/stop‑words or boiler‑plate (e.g., “The,” “Fact:”), yielding only superficial cues (Tab.~\ref{tab:failure-examples}). (\textit{ii}) \textbf{Nearby descriptive words or entity names.} In ConflictQA and Fakepedia (longer contexts), roughly half and about one‑third of failures, respectively, highlight adjacent proper nouns/adjectives (e.g., target entities or descriptors) while missing the gold token. (\textit{iii}) \textbf{Over‑attribution to question tokens.} Across datasets, HEs sometimes lock onto the interrogative phrase; this appears in about half of World‑Capital and ConflictQA failures and roughly one‑quarter of CounterFact and Fakepedia. Taken together, these patterns expose complementary weaknesses of HEs for pinpointing the exact answer location.

\textbf{Implications}
These patterns illustrate complementary failure modes for token‑level localisation: explanations can be distracted by high‑frequency tokens, gravitate to near‑answer descriptors, or focus on the question template rather than the evidence span. They highlight a real limitation: current HEs may fail to reliably pinpoint the precise evidence location even when the model copies the answer from context.

\begin{figure*}[p]
  \centering
  \begin{subfigure}[t]{0.48\textwidth}
    \centering
    \includegraphics[width=\textwidth]{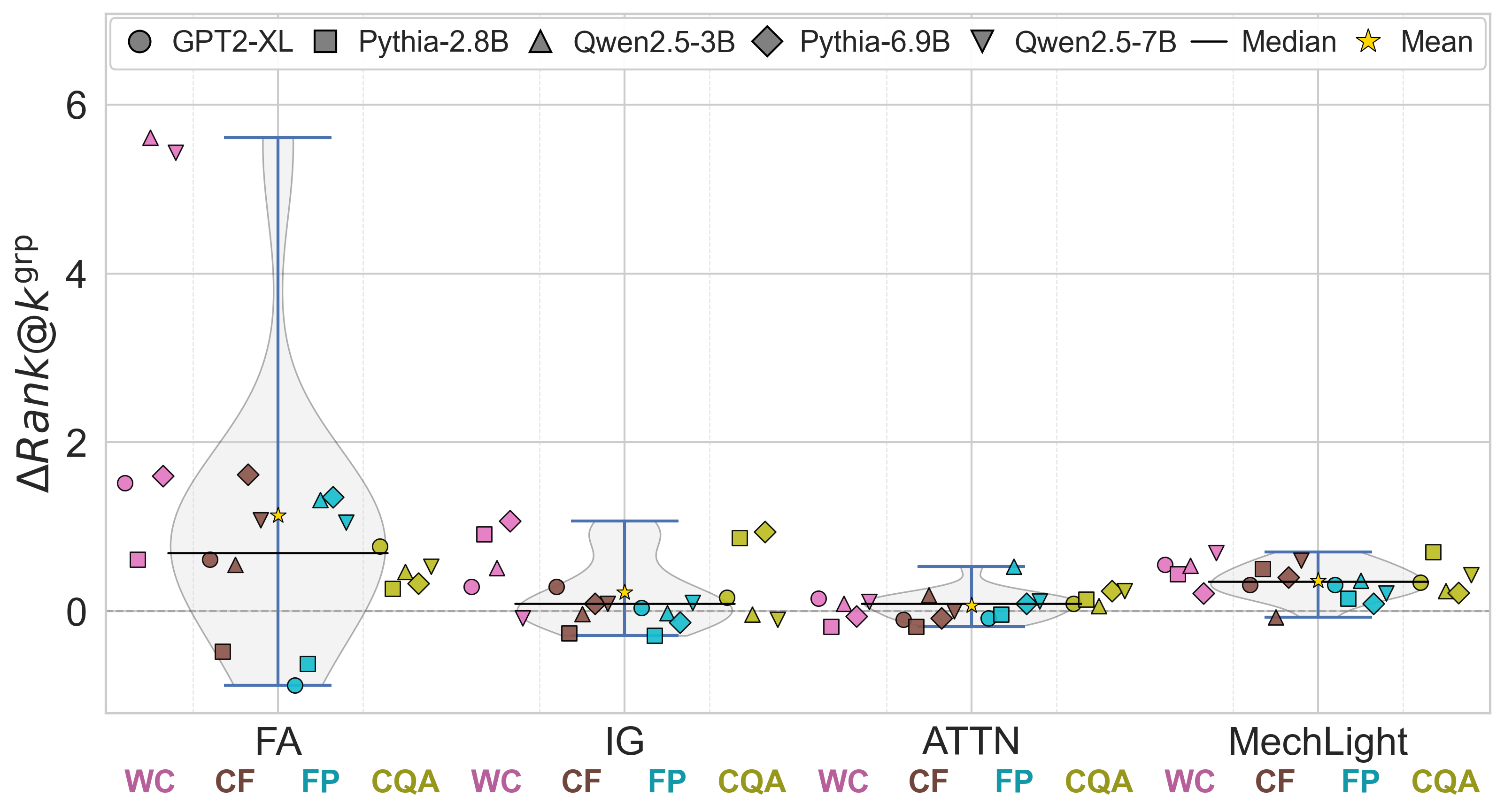}
    \caption{Top 3 Highlights in Conflicting Context}
  \label{fig:conflicting-topk-3-rank-datasets-margin-only}
  \end{subfigure}
  \hfill
  \begin{subfigure}[t]{0.48\textwidth}
    \centering
    \includegraphics[width=\textwidth]
{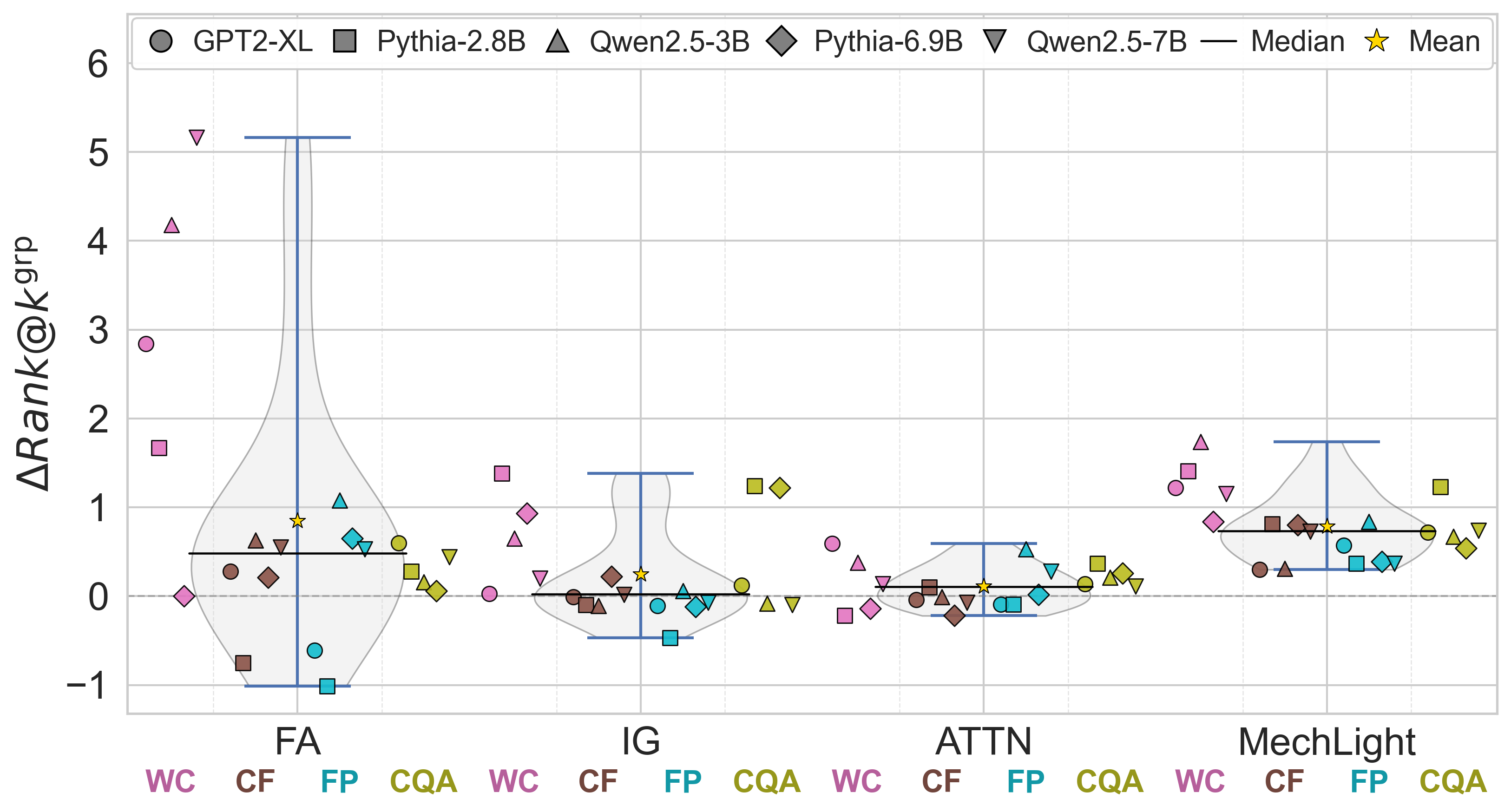}
    \caption{Top 9 Highlights in Conflicting Context}
  \label{fig:conflicting-topk-9-rank-datasets-margin-only}
  \end{subfigure}
    \caption{\DrankkGroup\ ($k=3;9$) (Eq. \ref{eq:drankk-cross-instance-group}) -- average margins for the explanation importance rank of context tokens in context vs. memory answer instances in \textbf{Conflicting} setup (\S\ref{sec:four-context-setups}). Positive and higher \DrankkGroup\ means the explanations can better distinguish the model's context usage behavior (choosing PK or CK for the answer (\S\ref{sec:rq1-discussion})). The colors denote the datasets, and the marker shapes denote the models.}
  
  \label{fig:conflicting-topk-3-and-topk-9-rank-datasets-margin-only}
\end{figure*}

\begin{figure*}[ht]
  \centering
  \begin{subfigure}[t]{0.48\textwidth}
    \centering
    \includegraphics[width=\textwidth]
    {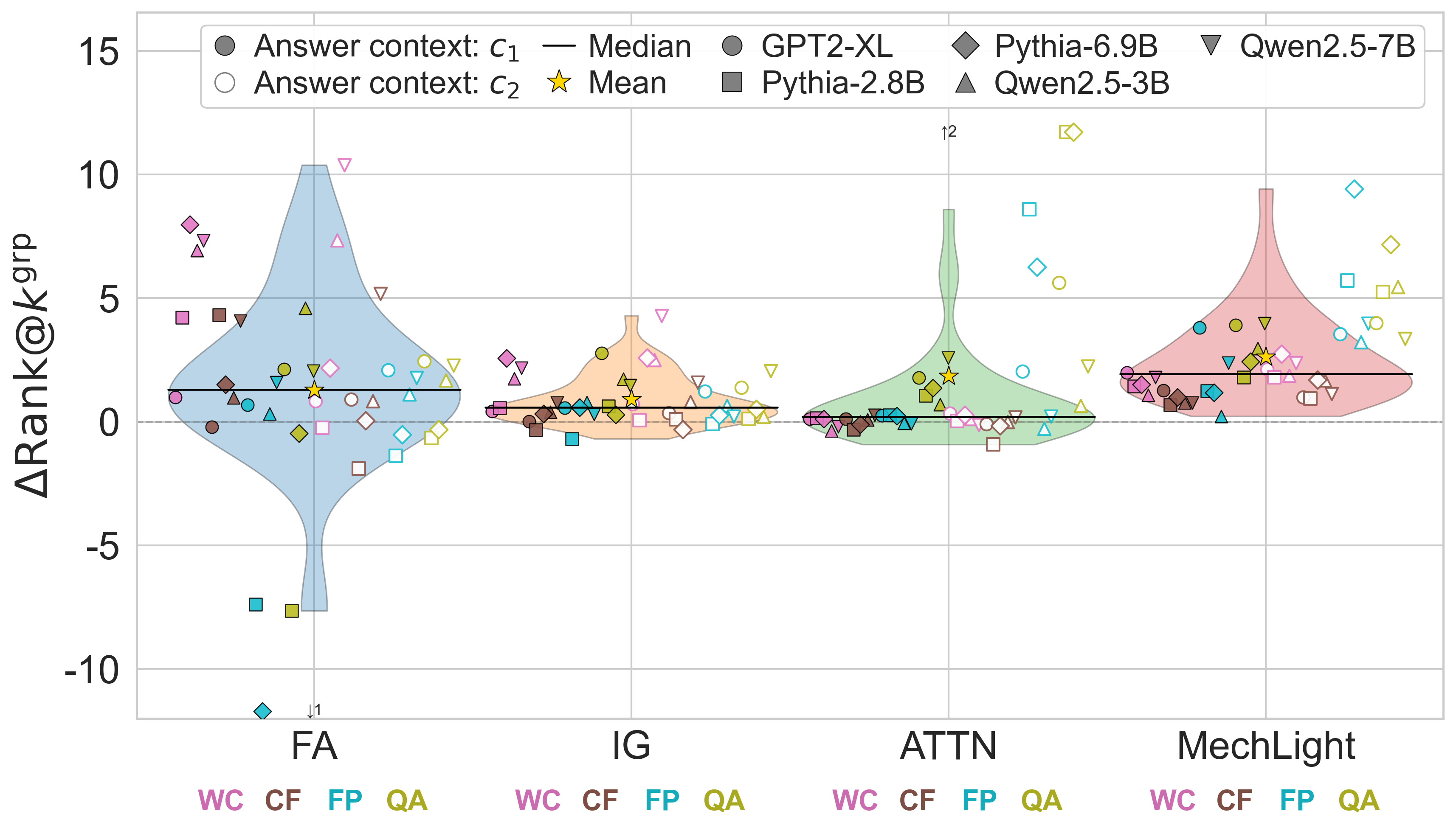}
    \caption{Top 3 Highlights in Double-Conflicting Contexts}
  \label{fig:double-conflicting-topk-3-rank-datasets-margin-only}
  \end{subfigure}
  \hfill
  \begin{subfigure}[t]{0.48\textwidth}
    \centering
\includegraphics[width=\textwidth]
{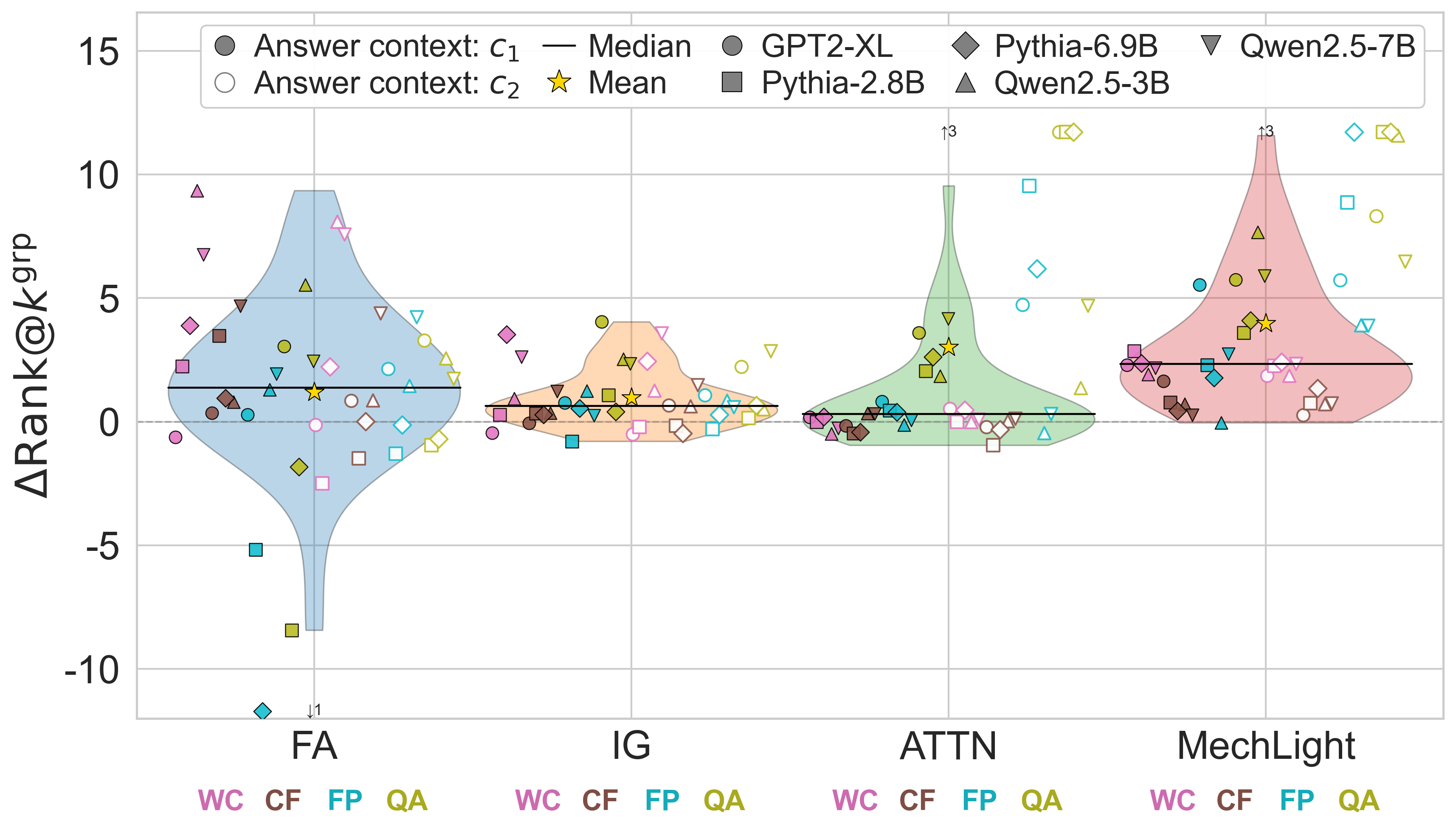}
    \caption{Top 9 Highlights in Double-Conflicting Contexts}
  \label{fig:double-conflicting-topk-9-rank-datasets-margin-only}
  \end{subfigure}
    \caption{\DrankkGroup\ ($k=3;9$) (Eq.~\ref{eq:drankk-cross-instance-group}) -- average margins for the rank of context $c_1$ and $c_2$ between two instance groups $D_{c_1}$ and $D_{c_2}$  in the \textbf{Double-Conflicting} setup (\S\ref{input-regime-2}). Higher \DrankkGroup is better, and a positive number means that the explanation can correctly indicate which context piece is used for the answer (\S\ref{sec:rq2-discussion}). The values are obtained by comparing the average importance scores of either the first context piece  ($c_1$, filled shapes at the left) or the second context piece ($c_2$, hollowed shapes at the right) between the two instance groups. The colors denote the datasets, and the marker shapes denote the models.
  }
  \label{fig:double-conflicting-topk-3-and-topk-9-rank-datasets-margin-only}
\end{figure*}

\begin{figure*}[ht]
  \centering
  \begin{subfigure}[t]{0.48\textwidth}
    \centering
    \includegraphics[width=\textwidth]
{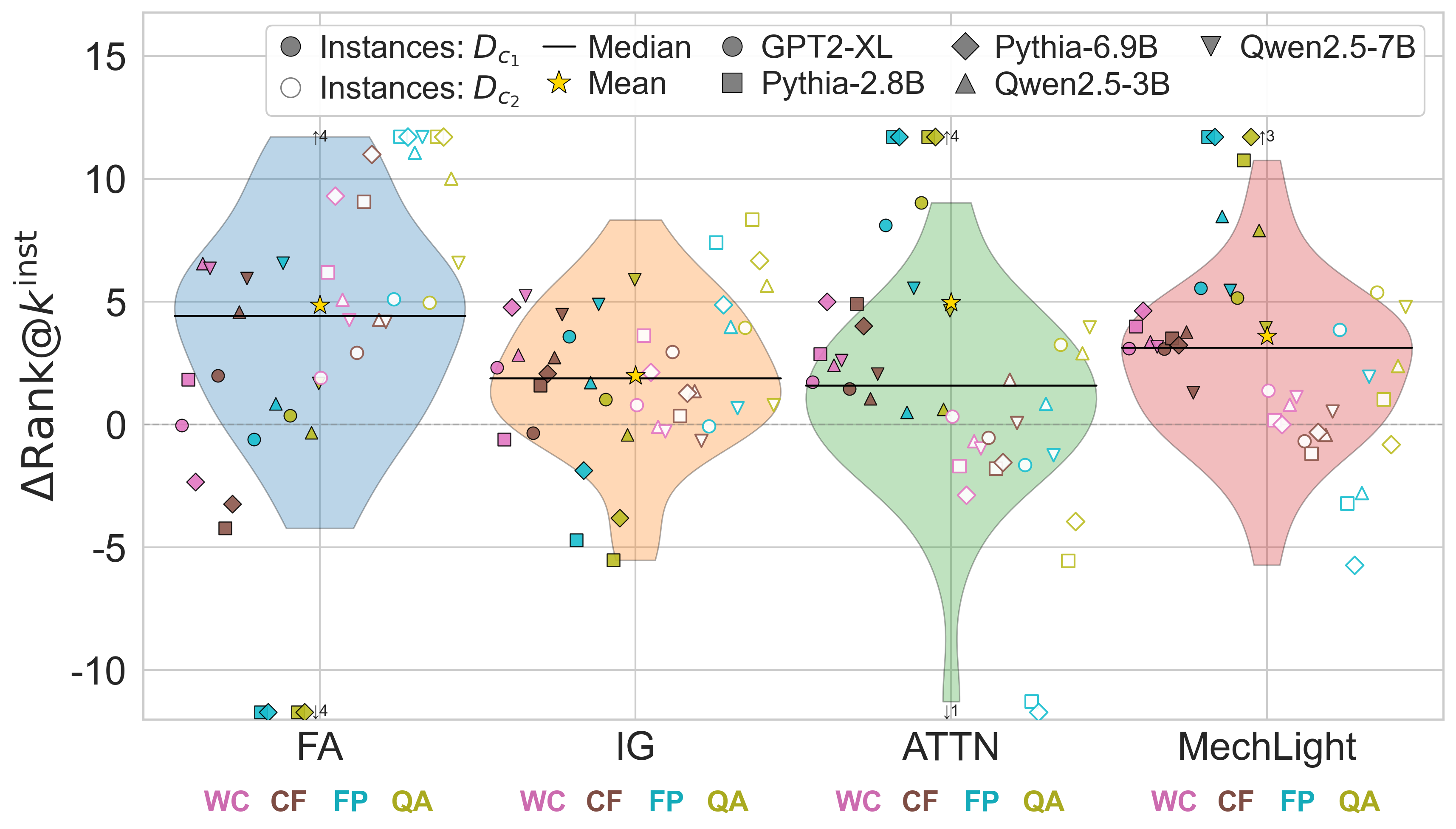}
    \caption{Top 3 Highlights in Double-Conflicting Contexts}
  \label{fig:double-conflicting-topk-3-rank-datasets-within-instance-margin}
  \end{subfigure}
  \hfill
  \begin{subfigure}[t]{0.48\textwidth}
    \centering
\includegraphics[width=\textwidth]
{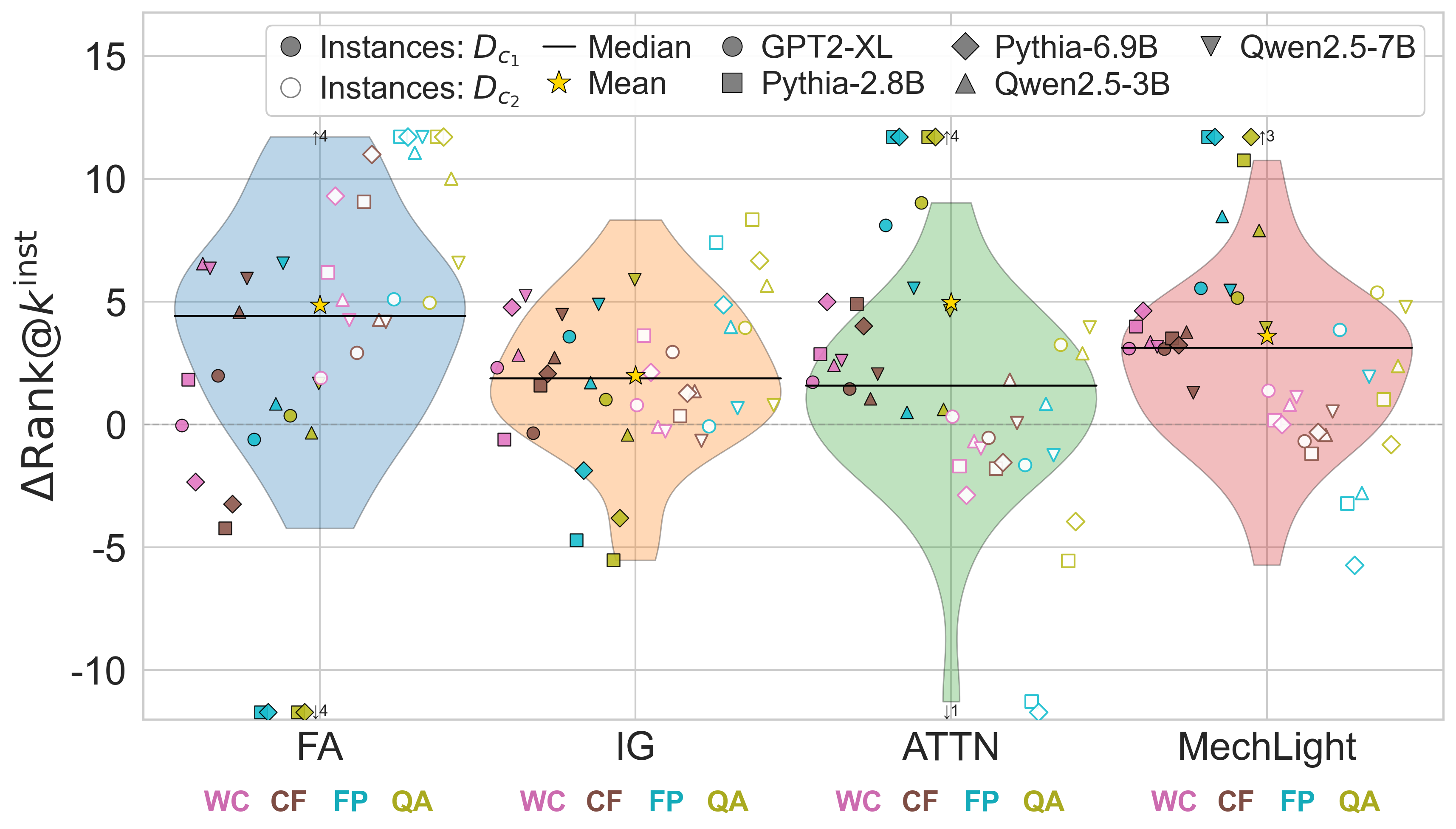}
    \caption{Top 9 Highlights in Double-Conflicting Contexts}
  \label{fig:double-conflicting-topk-9-rank-datasets-within-instance-margin}
  \end{subfigure}
  \caption{\DrankkInst\ ($k=3;9$)  (Eq.~\ref{eq:drankk-within-instance})  -- average \emph{within-instance-group} margins between the rank of the answer context piece and the other context piece in the \textbf{Double-Conflicting} setup (\S\ref{input-regime-2}).  Higher \DrankkInst\ is better, and a positive number means that the explanation can correctly indicate which context piece is used for the answer (\S\ref{sec:rq2-discussion}). The values are obtained by comparing importance scores of the answer-piece context and the other piece within either the instance group answering from the first context piece  ($D_1$, filled shapes at the left) or that answering from the second context piece ($D_1$, hollowed shapes at the right). The colors denote the datasets, and the marker shapes denote the models.}
  \label{fig:double-conflicting-topk-3-and-topk-9-rank-datasets-within-instance-margin}
\end{figure*}

\begin{figure*}[ht]
  \centering
  \begin{subfigure}[t]{0.48\textwidth}
    \centering
    \includegraphics[width=\textwidth]
    {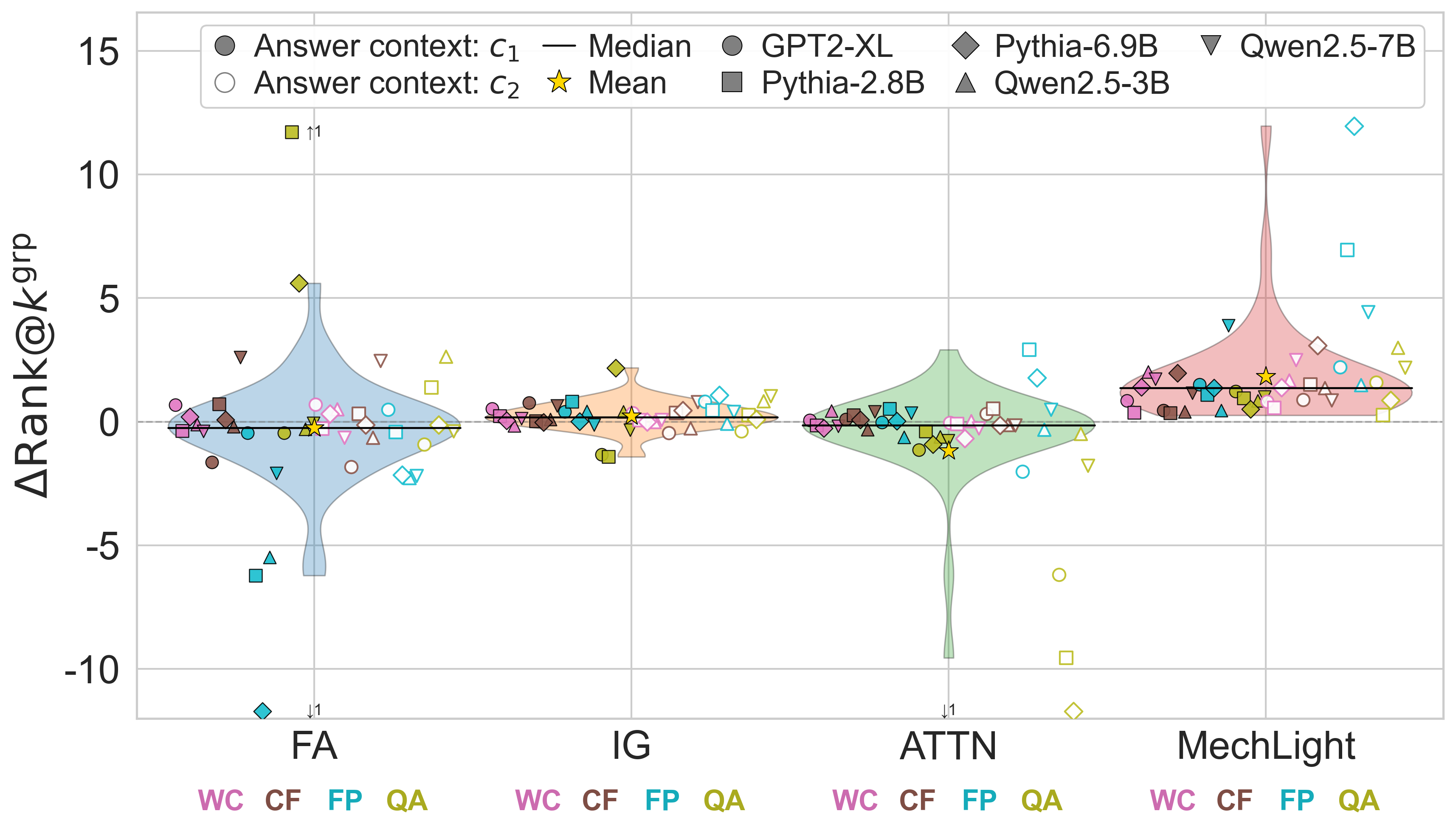}
    \caption{Double-Conflicting-Swap}
    \label{fig:conflicting-2-and-1-topk-5-rank-datasets-margin-only}
  \end{subfigure}
  \hfill
  \begin{subfigure}[t]{0.48\textwidth}
    \centering
    \includegraphics[width=\textwidth]
{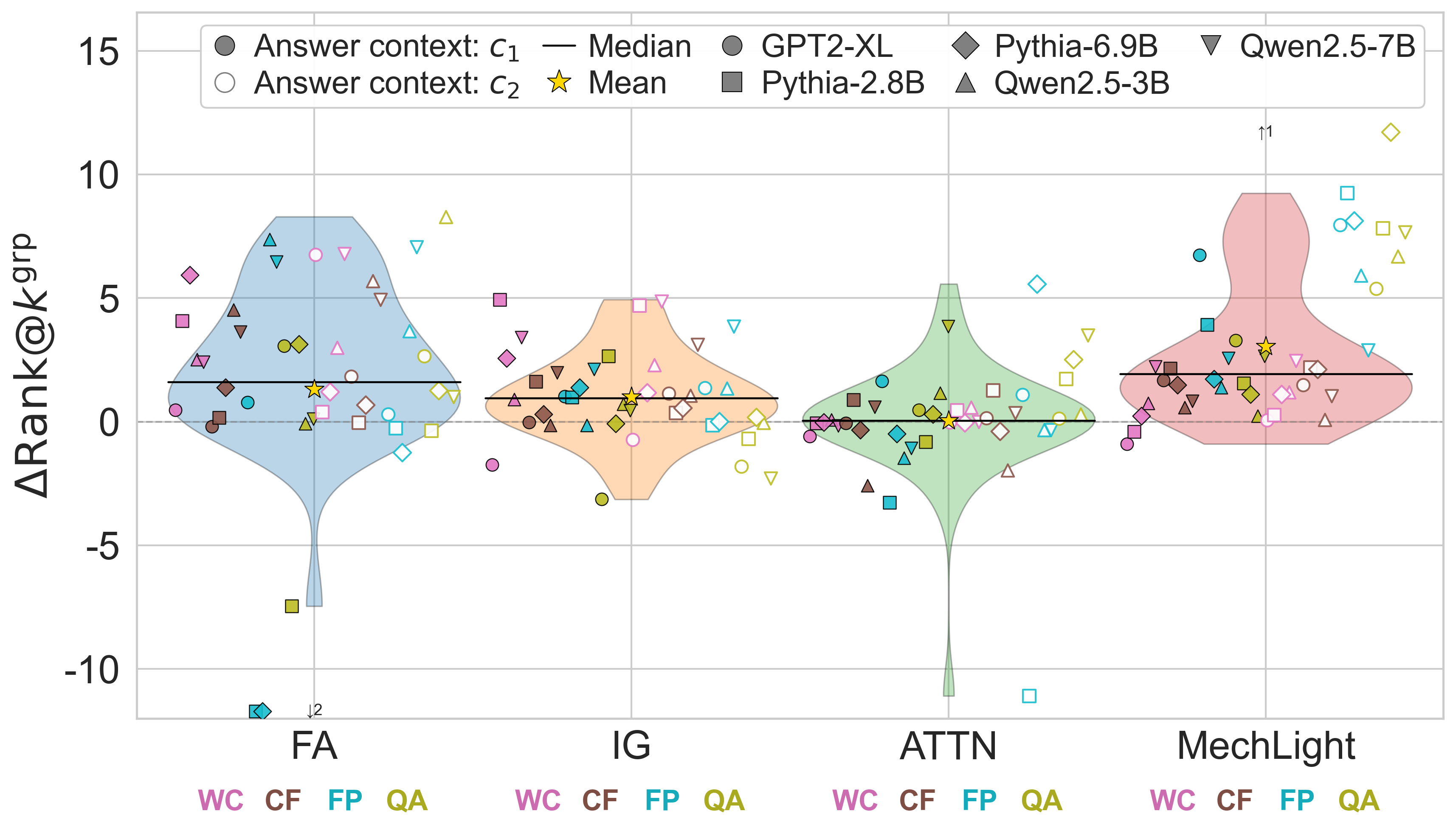}
    \caption{Mixed-Swap}
    \label{fig:conflicting-and-irrelevant-topk-5-rank-datasets-margin-only}
  \end{subfigure}
      \caption{\DrankkGroup\ (Eq.~\ref{eq:drankk-cross-instance-group}) -- average margins for the rank of context $c_1$ and $c_2$ between two instance groups $D_{c_1}$ and $D_{c_2}$  in the \textbf{Double-Conflicting} and \textbf{Mixed} setup \textbf{after swapping} the position of two context pieces (\S\ref{input-regime-2}). Higher \DrankkGroup is better, and a positive number means that the explanation can correctly indicate which context piece is used for the answer (\S\ref{sec:rq2-discussion}). The values are obtained by comparing the average importance scores of either the first context piece  ($c_1$, filled shapes at the left) or the second context piece ($c_2$, hollowed shapes at the right) between the two instance groups. The colors denote the datasets, and the marker shapes denote the models.
  }
  \label{fig:rank-topk-margin-double-conflicting-and-mixed-after-swap}
\end{figure*}

\begin{figure*}[ht]
  \centering
   \begin{subfigure}[t]{0.48\textwidth}
    \centering
    \includegraphics[width=\textwidth]
    {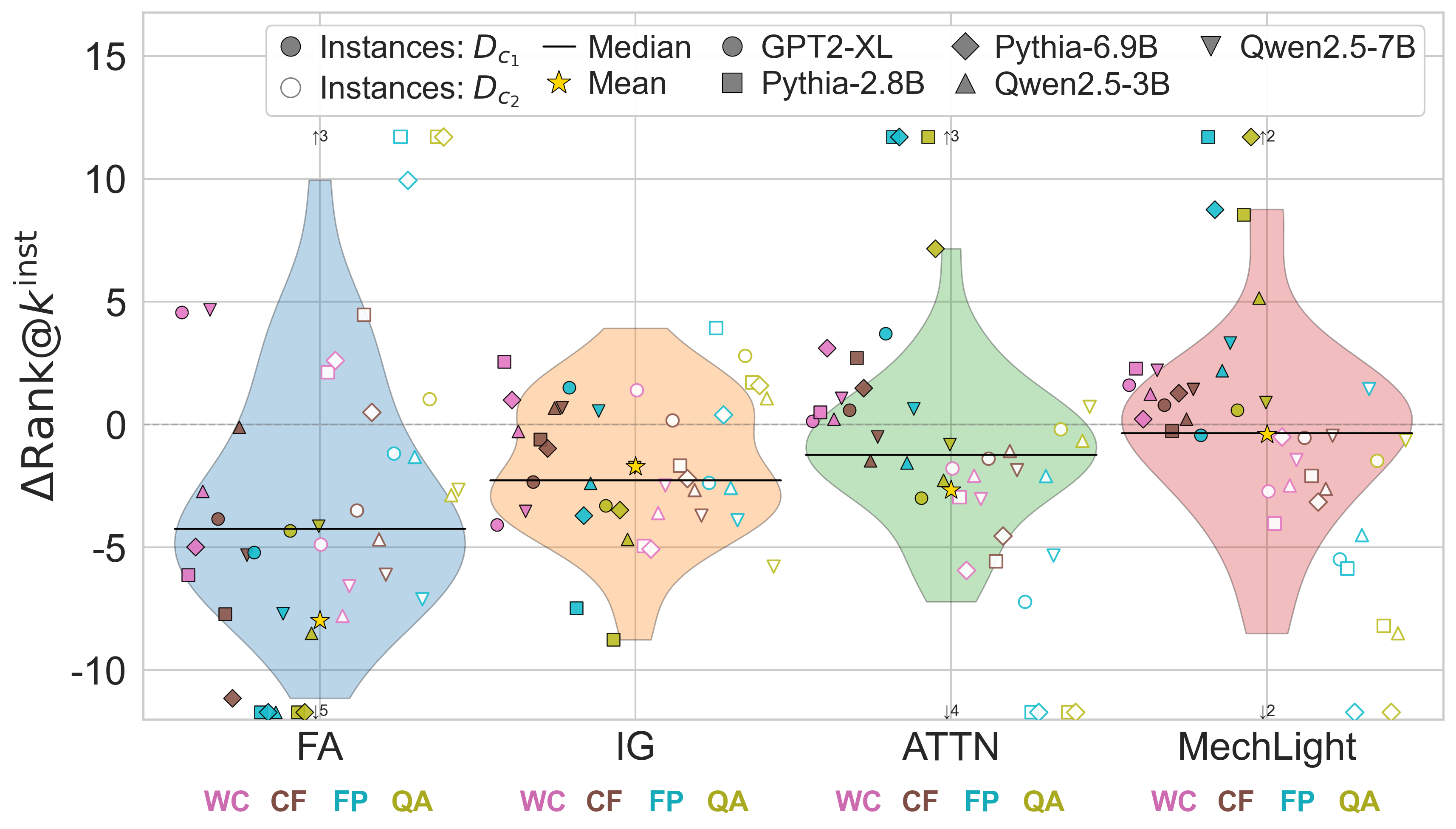}
    \caption{Double-Conflicting-Swap}
    \label{fig:conflicting-2-and-1-topk-5-rank-datasets-within-group-margin-only}
  \end{subfigure}
  \hfill
  \begin{subfigure}[t]{0.48\textwidth}
    \centering
    \includegraphics[width=\textwidth]{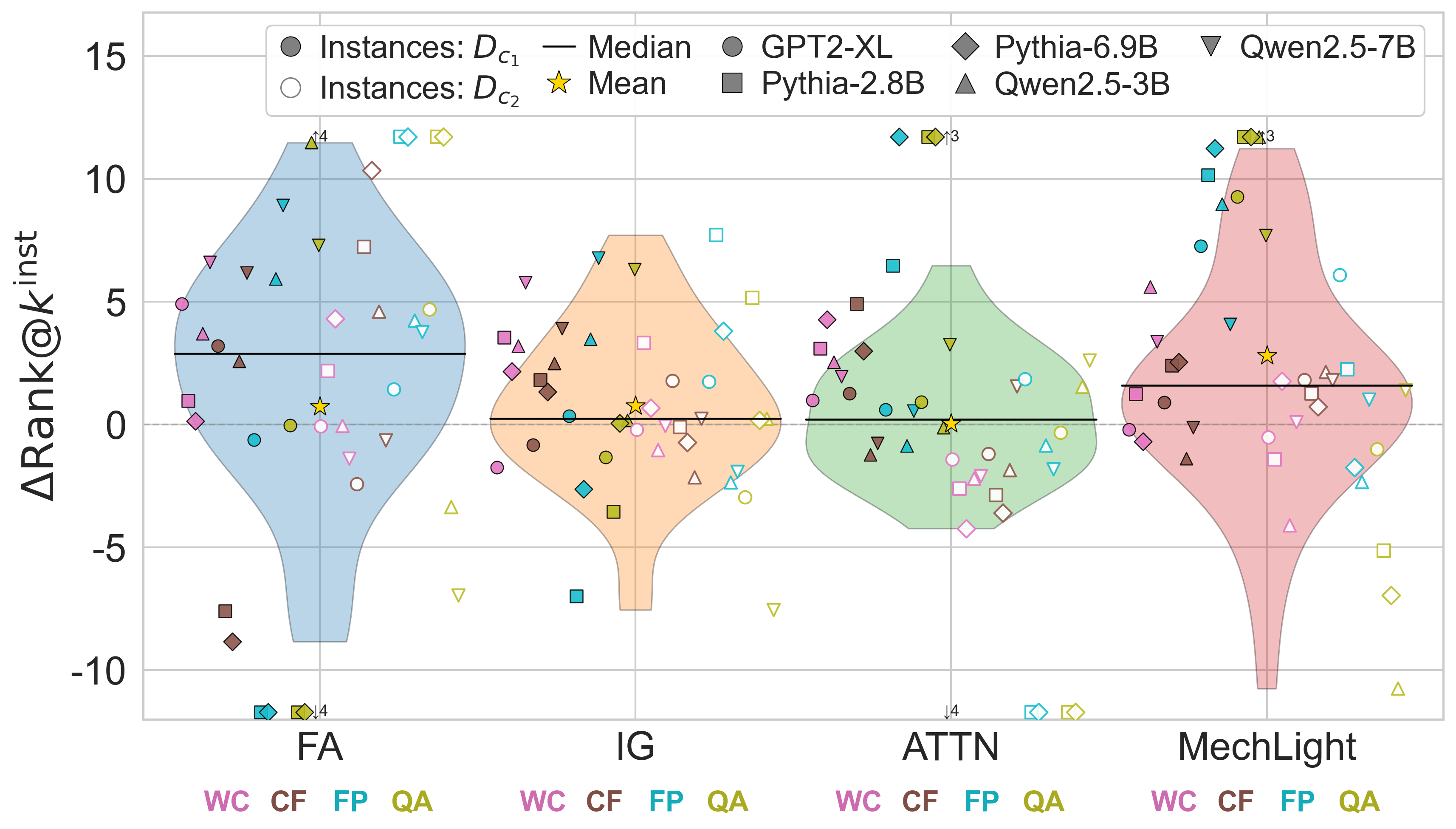}
    \caption{Mixed-Swap}
    \label{fig:conflicting-and-irrelevant-topk-5-rank-datasets-within-group-margin-only}
  \end{subfigure}
    \caption{\DrankkInst\ (Eq.~\ref{eq:drankk-within-instance})  -- average \emph{within-instance-group} margins between the rank of the answer context piece and the other context piece in the \textbf{Double-Conflicting} and \textbf{Mixed} setup after \textbf{swapping} the position of the two context pieces (\S\ref{input-regime-2}).  Higher \DrankkInst\ is better, and a positive number means that the explanation can correctly indicate which context piece is used for the answer (\S\ref{sec:rq2-discussion}). The values are obtained by comparing importance scores of the answer-piece context and the other piece within either the instance group answering from the first context piece  ($D_1$, filled shapes at the left) or that answering from the second context piece ($D_1$, hollowed shapes at the right). The colors denote the datasets, and the marker shapes denote the models.}
  \label{fig:rank-topk-margin-within-instance-double-conflicting-and-mixed-after-swap}
\end{figure*}


\begin{figure*}[ht]
  \centering
  \begin{subfigure}[t]{0.48\textwidth}
    \centering
    \includegraphics[width=\textwidth]
    {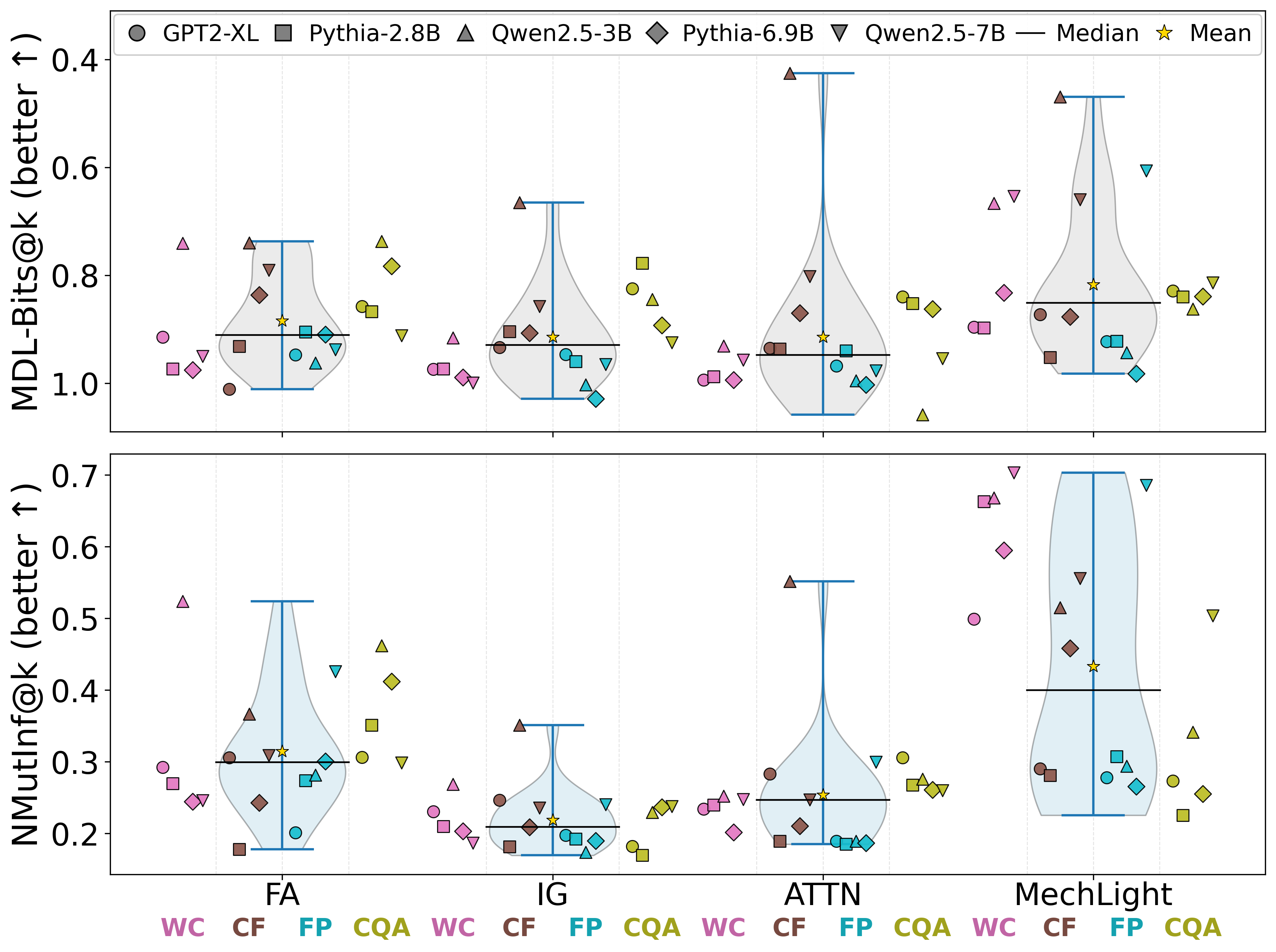}
    \caption{Double-Conflicting-Swap}
    \label{fig:conflicting-2-and-1-topk-5-mutual-information-and-mdl-prob}
  \end{subfigure}
  \hfill
  \begin{subfigure}[t]{0.48\textwidth}
    \centering
    \includegraphics[width=\textwidth]
{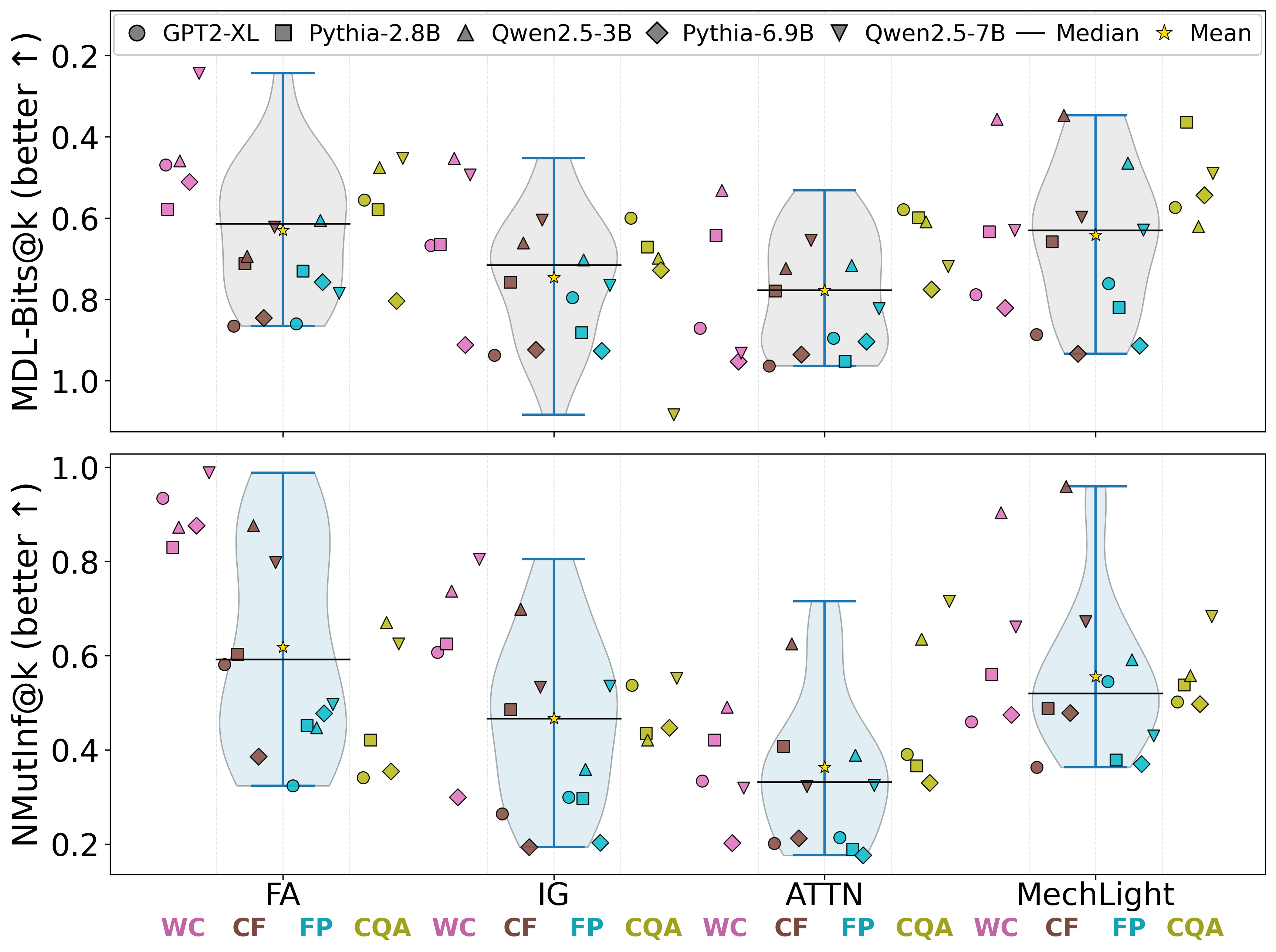}
    \caption{Mixed-Swap}
    \label{fig:conflicting-and-irrelevant-topk-5-mutual-information-and-mdl-prob}
  \end{subfigure}
  \caption{\MDLBitsk\ (top; y-axis is inverted; Eq.~\ref{eq:mdl}) and \NMutInfk\ (bottom; Eq.~\ref{eq:mutual-information}) in \textbf{Double-Conflicting} and \textbf{Mixed} setups after \textbf{swapping} the position of the two context pieces (\S\ref{sec:four-context-setups}). Both \MDLBitsk (smaller value) and \NMutInfk are higher the better, indicating that the explanation importance scores are better at revealing the model's answer selection across the two context passages, i.e., from the first vs from the second piece. (\S\ref{sec:rq2-discussion}). The colors denote the datasets, and the marker shapes denote the models.
  }
  \label{fig:topk-mutual-information-and-mdl-prob-double-conflicting-or-mixed-swapped}
\end{figure*}

\begin{table*}[t]
\centering
\small
\setlength{\tabcolsep}{3pt}
\renewcommand{\arraystretch}{0.95}
\begin{tabular}{@{}ccccc|cc@{}}
\toprule
& & & \multicolumn{2}{c|}{\textbf{Conflicting}} & \multicolumn{2}{c}{\textbf{Double-Conflicting}} \\
\cmidrule(lr){4-5} \cmidrule(lr){6-7}
\textbf{Dataset} & \textbf{Model} & \textbf{Method} & AOPC$_\text{comp}\!\uparrow$ & AOPC$_\text{suff}\!\downarrow$ & AOPC$_\text{comp}\!\uparrow$ & AOPC$_\text{suff}\!\downarrow$ \\
\midrule
\multirow{8}{*}{WorldCapital}
& \multirow{4}{*}{Qwen2.5-7B}
  & \FA   & 122.7 & 150.61 & 182.7 & 250.98 \\
& & \IG   & 118.0 & \bestsuff{149.33} & 180.7 & 255.75 \\
& & \Attn & \bestcomp{127.9} & 153.76 & \bestcomp{184.4} & 245.35 \\
& & \MI   & 119.3 & 151.61 & 177.3 & \bestsuff{244.78} \\
\cmidrule(lr){2-7}
& \multirow{4}{*}{Pythia-6.9B}
  & \FA   & 113.6 & \bestsuff{147.09} & \bestcomp{187.7} & 267.13 \\
& & \IG   & \bestcomp{114.5} & 147.82 & 186.7 & 267.90 \\
& & \Attn & 96.4  & 151.13 & 155.4 & 267.18 \\
& & \MI   & 104.2 & 151.90 & 174.8 & \bestsuff{265.56} \\
\midrule
\multirow{8}{*}{Fakepedia}
& \multirow{4}{*}{Qwen2.5-7B}
  & \FA   & 843.7 & 1047.21 & \bestcomp{1690.2} & 2294.96 \\
& & \IG   & 834.6 & 1052.99 & 1688.5 & 2283.39 \\
& & \Attn & \bestcomp{849.1} & 1028.22 & 1645.8 & 2340.83 \\
& & \MI   & 819.0 & \bestsuff{1019.60} & 1621.8 & \bestsuff{2283.02} \\
\cmidrule(lr){2-7}
& \multirow{4}{*}{Pythia-6.9B}
  & \FA   & \bestcomp{815.6} & \bestsuff{1100.21} & 1683.5 & \bestsuff{2482.10} \\
& & \IG   & 811.1 & 1111.49 & \bestcomp{1685.4} & 2499.73 \\
& & \Attn & 614.1 & 1115.60 & 817.8  & 2493.53 \\
& & \MI   & 691.1 & 1108.89 & 1023.4 & 2500.70 \\
\bottomrule
\end{tabular}
\caption{Faithfulness in \textbf{Conflicting} and  \textbf{Double-Conflicting} contexts for two models on two datasets (short-context: World Capital, long-context: Fakepedia). Higher AOPC$_\text{comp}$, lower AOPC$_\text{suff}$ is better. Best entries are \bestcomp{underlined} for comprehensiveness and \bestsuff{bold} for sufficiency. \MI\ and \FA\ are the best AOPC$_\text{suff}$; \Attn\ is the best for AOPC$_\text{comp}$ in Conflicting setup.}
\label{tab:faithfulness}
\end{table*}

\begin{figure*}[p]
  \centering
  \includegraphics[width=\columnwidth]{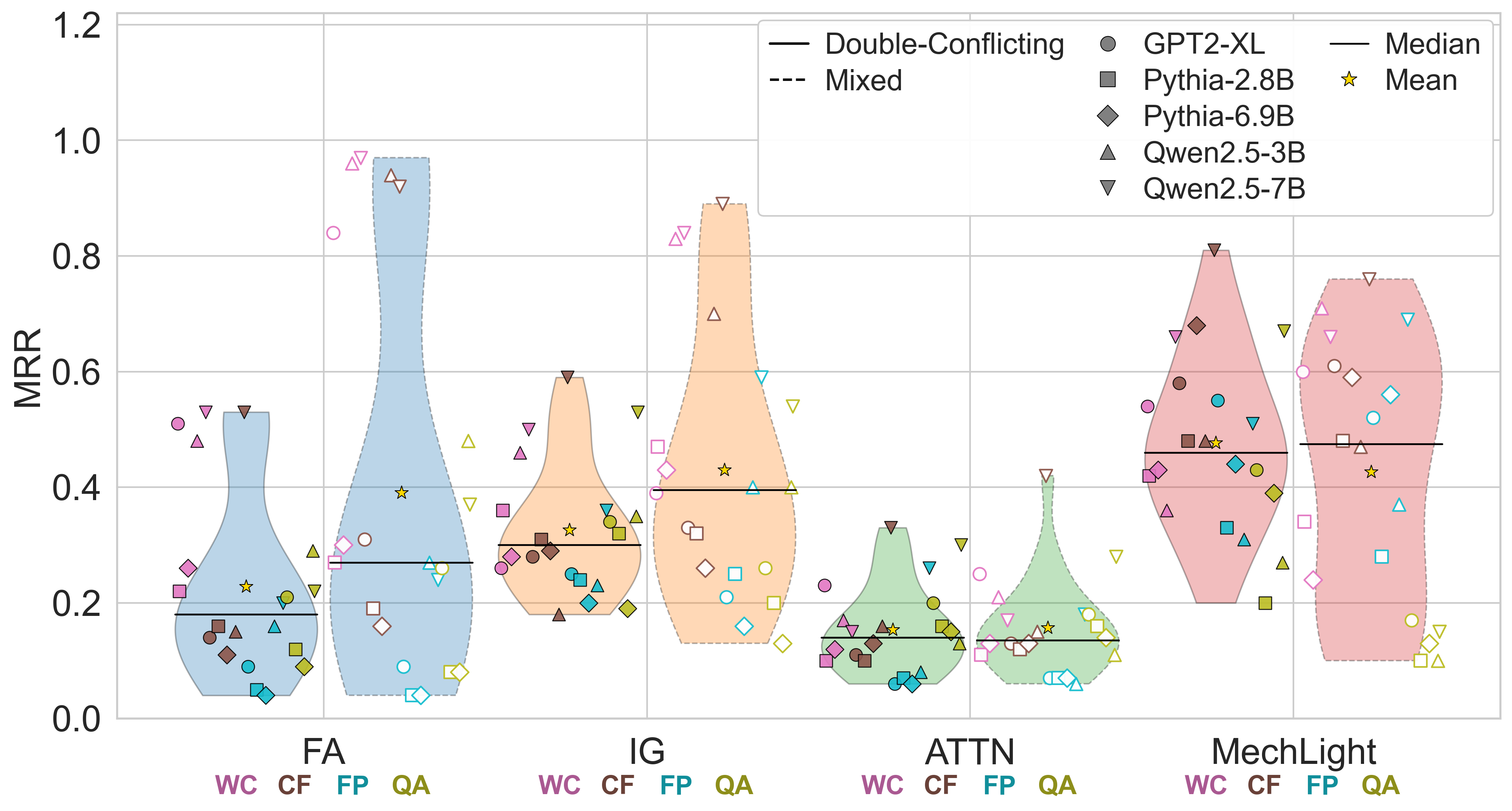}
    \caption{\mrr\ (Eq.~\ref{eq:mrr}) -- Mean Reciprocal Rank for the predicted answer tokens within the context-answer instances for \textbf{Double-Conflicting} and \textbf{Mixed} setups after \textbf{swapping} the position of the two context pieces (\S\ref{sec:four-context-setups}). Higher \mrr\ is better; larger values indicate the true answer token is placed near the top of the ranked list, meaning the explanation can more precisely indicate the answer location (\S\ref{sec:rq3-discussion}). Filled shapes stand for \textbf{Conflicting} setup and \textbf{Double-Conflicting} setup in the left and right subplot, respectively; hollow shapes stand for \textbf{Irrelevant} setup and \textbf{Mixed} setup in the left and right subplot, respectively. The colors denote the datasets, and the marker shapes denote the models.
  }
  \label{fig:answer-token-mrr-double-or-mixed-after-swap}
\end{figure*}

\end{document}